\documentclass[acmlarge,anonymous=false]{acmart}
\usepackage{booktabs}
\usepackage[ruled,vlined]{algorithm2e}
\usepackage{multirow}
\usepackage{subcaption}

\AtBeginDocument{%
  \providecommand\BibTeX{{%
    \normalfont B\kern-0.5em{\scshape i\kern-0.25em b}\kern-0.8em\TeX}}}

\setcopyright{acmlicensed}
\acmJournal{IMWUT}
\acmYear{2022} \acmVolume{6} \acmNumber{1} \acmArticle{28} \acmMonth{3} \acmPrice{15.00}\acmDOI{10.1145/3517252}

\begin{document}

\title{Learning Disentangled Behaviour Patterns for Wearable-based Human Activity Recognition}

\author{Jie Su}
\orcid{0000-0002-1427-1253}
\affiliation{%
  \institution{Newcastle University}
  \city{Newcastle Upon Tyne}
  \country{United Kingdom}
}
\email{jieamsu@gmail.com}

\author{Zhenyu Wen}
\orcid{0000-0002-2914-912X}
\affiliation{%
  \institution{Zhejiang University of Technology}
  \city{Hangzhou}
  \country{China}
  }
\email{wenluke427@gmail.com}

\author{Tao Lin}
\orcid{0000-0002-3246-6935}
\affiliation{%
  \institution{EPFL}
  \city{Lausanne}
  \country{Switzerland}
}
\email{tao.lin@epfl.ch}

\author{Yu Guan}
\orcid{0000-0002-1283-3806}
\affiliation{%
  \institution{Newcastle University}
  \city{Newcastle Upon Tyne (Corresponding Author)}
  \country{United Kingdom}
}
\email{yu.guan@newcastle.ac.uk}

\renewcommand{\shortauthors}{Su et al.}

\begin{abstract}
In wearable-based human activity recognition (HAR) research, one of the major challenges is the large intra-class variability problem. The collected activity signal is often, if not always, coupled with noises or bias caused by personal, environmental, or other factors, making it difficult to learn effective features for HAR tasks, especially when with inadequate data. To address this issue, in this work, we proposed a Behaviour Pattern Disentanglement (BPD) framework, which can disentangle the behavior patterns from the irrelevant noises such as personal styles or environmental noises, etc. Based on a disentanglement network, we designed several loss functions and used an adversarial training strategy for optimization, which can disentangle activity signals from the irrelevant noises with the least dependency (between them) in the feature space. Our BPD framework is flexible, and it can be used on top of existing deep learning (DL) approaches for feature refinement. Extensive experiments were conducted on four public HAR datasets, and the promising results of our proposed BPD scheme suggest its flexibility and effectiveness. This is an open-source project, and the code can be found at http://github.com/Jie-su/BPD
\end{abstract}



\maketitle

\section{Introduction}
Wearable-based HAR is one of the most popular themes in ubiquitous and wearable computing, and it plays a major role in a wide range of applications such as health assessment~\cite{Autism, babystroke}, sleeps monitoring~\cite{bing2020}, sports coaching~\cite{Cricket}, etc.  The main tasks of wearable-based HAR involve partitioning the multi-variate data stream from one or more sensors into segments and assigning a corresponding activity label to each segment~\cite{qian2019novel}.

Previous studies in this field leveraged the hand-crafted features in statistical~(e.g., mean, variance) and frequency~(e.g., power spectral density) domain to represent segments of raw sensory streams and projected the feature vector to the corresponding activity labels based on traditional machine learning methods such as SVM~\cite{Cricket}, KNN~\cite{Autism} and Random Forest~\cite{pal2005random}. However, designing effective features tends to be a trial-and-error process, and discriminant features may vary from task to task, making system-developing expensive and less sustainable. To solve this problem, recent studies~\cite{2015MCCNN,2016bLSTMS,2016DeepConvLSTM,2017ensembleslstm, 2018attentionlstm} leveraged the exceptional data representation ability of deep learning methods to expedite feature extraction. Such studies mainly utilized the deep neural networks~(e.g., Convolutions neural networks(CNN)~\cite{lecun1995convolutional}, Long-Short Term Memory(LSTM)~\cite{hochreiter1997long}) to extract the features from the original input sensors in an end-to-end manner.   

\begin{figure*}[t]
    \centering
    \includegraphics[width=\textwidth]{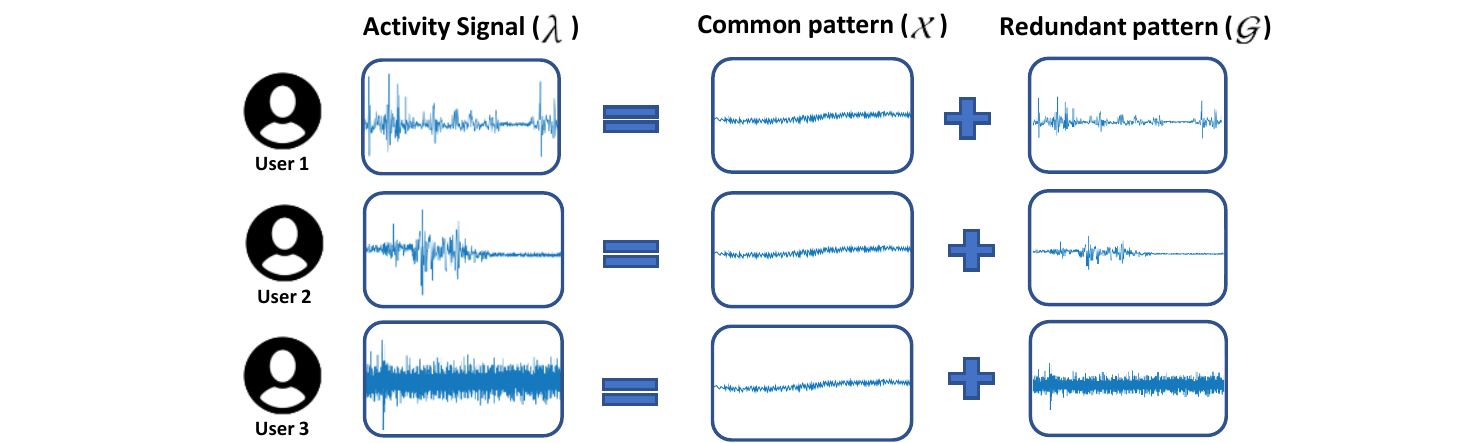}
    
    \caption{Disentanglement factors in the feature representation for activity signal data: performing activity recognition over the disentangled features ($\mathcal{X}$) is much less challenging than that of raw sensor data ($\lambda$). 
  The first column presents the raw sensor data ($\lambda$) for the standing activity across different user groups. The second and third columns indicate the disentangled common pattern ($\mathcal{X}$) and redundant/irrelevant patterns ($\mathcal{G}$)~(e.g., gender, physical strength, etc.), respectively. }
    \label{fig:intro_exp}
\end{figure*}

Although the deep learning approaches can extract decent representation from input sensor data, they may face challenges when dealing with multi-model sensor streams from diverse subjects/users. One of the crucial challenges is the intra-class variability problem. The discrepancy between subjects in performing activities was neglected in current studies - they usually map all subjects indiscriminately to the high-level feature representations with Deep Learning methods. However, the sensor record for the same activity may vary among different people due to their personal characteristics, such as gender, habits, physical strength, etc. Figure \ref{fig:intro_exp} shows the sensor reading of the standing activity $\lambda=\mathcal{X}+\mathcal{G}$. $\mathcal{X}$ represents standing activity which is the common pattern (distribution) across various groups of users.  $\mathcal{G}$ are the \emph{redundant} patterns which vary between different group of users, even each independent user. Thus, such redundant patterns bring challenges for developing a robust activity recognition system, serving million or billion of users. To overcome this, we can collect the new sensor data and retrain the model to increase the generalisation ability. However, this solution is time-consuming and it is very costly to label the new sensor data. Alternatively, removing or disentangling the redundant patterns from the sensor data can significantly improve robustness and generalisation of activity recognition system. 

Recently, learning disentangled representations has attracted a lot of attention from the machine learning community. Such representations provide many advantages: improving the predictive performance on downstream tasks~\cite{locatello2019challenging,locatello2019disentangling}, reducing the sample complexity~\cite{ridgeway2018learning,van2019disentangled}, offering interpretability~\cite{higgins2016beta}, improving fairness~\cite{locatello2019fairness} and have been identified as a way to overcome \textit{short cut} learning in deep learning~\cite{geirhos2020shortcut}. From the literature, disentanglement learning approaches proved to be effective in the computer vision field. However, applying disentanglement to sensor data~(e.g., human activity recognition) is more challenging since the disentanglement should consider both context level~(e.g., time dimension) information and feature level information.

To address this issue, in this work we proposed a Behaviour Pattern Disentanglement (BPD) scheme, which utilizes disentanglers to induce two groups of representations. Ideally, the activity features are captured as the common patterns on a certain class of activity, and the redundant representations are captured as the unpredictable personal patterns such as the lifestyle of a person. To effectively disentangle two groups of features, we develop an adversarial disentangle mechanism. By using such mechanism, the generated activity feature representations are expected to be more invariant to other domains, compared to the original data. Moreover, our BPD framework is flexible, and is applicable on top of existing popular DL approaches, such as CNN~\cite{2015MCCNN}, DeepConvLSTM~\cite{2016DeepConvLSTM}, etc. for activity feature refinement. To evaluate our models, we leveraged the Leave-one-subject-out cross validation (LOSO-CV) protocol on the four public HAR datasets, which can demonstrate the performance at both overall and the subject-level. Our contributions can be summarised as follows:
\begin{itemize}
\item We proposed the BPD framework, which can separate the activity signal from the redundant feature in the feature space with the least dependency.  
\item Our BPD scheme is flexible, and it can be used on top of existing DL approaches for feature refinement, with improved HAR results. Our project is open source and the code can be found at http://github.com/Jie-su/BPD
\item Extensive experiments were conducted, and we studied our BPD framework in details. The promising results suggested its effectiveness.
\end{itemize}

The rest of this paper is organized as follow. Section 2 introduces the related background knowledge. Section 3\&4 present the problem definition and the details of the proposed BPD framework. Section 5 gives the experimental settings as well as evaluation results, and Section 6 concludes.

\section{Background}
 HAR has a long-standing history in the wider ubiquitous and wearable computing community. Recently, a multitude of methods have been proposed and facilitate a variety of applications. HAR has become one of the pillars of the third generation of computing~\cite{schmidt1999there}. In the following section, we will review the specific background for this paper, which spans three main subject areas: i) Deep learning for HAR in ubiquitous and wearable computing; ii) Adversarial Learning; iii) Representation Disentangle Learning.

\subsection{Human Activity Recognition}
Traditional machine learning\cite{qian2020orchestrating} approaches such as K-Nearest Neighbor (KNN), Hidden Markov Model (HMM), Support Vector Machine (SVM), Random Forest (RF), and Naive Bayes (NB) have been successfully applied on HAR~\cite{bao2004activity, kwapisz2011activity, guo2016wearable}. The main drawback of these models is that they are mainly relying on hand-crafted features or heuristic information.

Deep learning methods can automatically extract features from raw signals, reducing the efforts on feature engineering procedures. 
One of the most popular deep learning model is convolutional neural network (CNN), which can 
extract the HAR representation by stacking multiple convolutional layers \cite{2015MCCNN}. DeepConvLSTM~\cite{2016DeepConvLSTM} extended CNN by adding LSTM layers for temporal information modelling. In~\cite{hammerla2016deep}, Hammerla \textit{et al.} comprehensively studied the performance of DNNs, CNNs and RNNs for HAR tasks. Guan and Ploetz ~\cite{2017ensembleslstm} explored sample-wise activity recognition by ensembles of deep LSTM learners using an epoch-wise bagging scheme. Murahari and Ploetz~\cite{2018attentionlstm} added the attention layers to the DeepConvLSTM model to learn local temporal context from raw sensor data. Recent work~\cite{2019DDNN} proposed a DDNN model to learn statistical, temporal and spatial correlation features from signals, before a final fusion for performing the activity recognition. 

\subsection{Adversarial Learning}
As deep neural networks have found their way from labs to the real world, the security and integrity of the applications pose a great concern. Adversaries can craftily manipulate legitimate inputs, which may be imperceptible to the human eye, but can force a trained model to produce incorrect outputs~\cite{chakraborty2018adversarial}. Szegedy \textit{et al}.~\cite{szegedy2013intriguing} first discovered that well-trained deep neural networks were susceptible to adversarial attacks. Attacks on autonomous vehicles have been demonstrated by Kurakin \textit{et al}. where the adversary manipulated traffic signs successfully confuse the learning model. 

The Generative Adversarial Networks~(GANs) were proposed by Goodfellow \textit{et al}.~\cite{goodfellow2014generative} which brought the concept of adversarial to the network level. More precisely, the key idea of GANs is to create a competition between the generative model and an adversary: a discriminator model that learns to determine whether a sample is from the model distribution or the data distribution~\cite{goodfellow2014generative}. Imagining the generative model as a team of counterfeiters trying to produce fake currency that is non-detectable, while the discriminator as the police trying to detect the counterfeit currency. Such competition drives both teams to improve their intelligence until the counterfeits are indistinguishable from the genuine currency. Benefiting from GANs, the concept of adversarial learning has become a popular research topic in the deep learning community and is applied to many applications such as adversarial sample generation~\cite{xiao2018generating}, style transfer~\cite{jing2019neural}, and autopilot~\cite{zhang2018deeproad}.

\subsection{Representation Disentangle Learning}
Prior to the deep learning era, most computer vision systems made use of features that were hand-engineered and task-oriented. One of the desired goals and challenges for these features was to be invariant to certain nuisance/redundant factors in the data such as affine transforms, blur, etc. Early studies such as Gopalan \textit{et al.} \cite{gopalan2012blur} and  Lowe  \cite{lowe2004distinctive} have achieved it, but the drawback of these methods is that they are mainly relying on hand-crafted features. Recent advanced deep learning techniques are primarily data-driven where features are learned by adding suitable constraints on the learning paradigm. Being dependent on data enables those methods to learn covariate factors in the data~(e.g., angle, shape, the noise in the data generating process). It should be noted that the `noise' can be any undesired and unknown factors of original data which we parameterise with a mathematical model. Figure~\ref{fig:intro_exp} illustrates the concept of covariate factors that might exist in the feature representation of multi-dimensional signals~(i.e., time-series signal). Specifically, it illustrates common activity factor and personal/environmental factors for time-series data. 

Disentangling those factors can help us to further explore the highly entangled high dimensional data, but the factors might become the bias/noise to the recognition system. There are a few common kinds of nuisance/noise factors that creep into the sensor signal datasets which can be used for training recognition systems: gender, physical strength variation, age, etc. Here, the nuisance/noise factors can be defined as `task-based undesired factors of variations' since certain factors of variation are desirable for some tasks while not undesirable for others. For example, gender could be a noise factor when doing activity classification but could be a key factor for conditional signal sample generation~(i.e., generate signal samples with gender condition). Thus, exploring the disentanglement and disentangling desired representation/factors is crucial for many downstream machine learning applications.   

Benefiting from the advance of DL techniques, recent works~\cite{mathieu2016disentangling, liu2018unified, odena2017conditional, hu2018dual, hu2020robust,hu2018robust,wang2019order} in computer vision started to learn the interpretable representations from images or videos by utilising generative adversarial networks. (GANs)~\cite{goodfellow2014generative} or Variational autoencoders(VAEs)~\cite{kingma2013auto}. InfoGAN~\cite{chen2016infogan} was proposed to learn the disentangled representation in an unsupervised manner while it may suffer from training instabilities. Beta-VAE~\cite{higgins2016beta} improved the poor disentanglement/reconstruction trade-off of the original VAEs. Later, Liu et al.~\cite{liu2018unified} introduced a unified feature disentanglement framework to learn domain invariant features across different domains. Recently, disentangled representation learning has also been applied to some popular applications (e.g., Gait recognition~\cite{hu2020robust}, speaker recognition~\cite{sang2020deaan}). Hu et al.~\cite{hu2020robust} proposed a disentanglement framework that can separate gait identity from the camera view for view-invariant gait recognition. DEAAN~\cite{sang2020deaan} was proposed to disentangle speaker-related features from speech signals to achieve robust speaker adaptation and recognition. Recently, GILE~\cite{2021GILE} was proposed by Qian et al. to disentangle ID information from raw sensor data. They utilised the additional subject ID label and the Independence Excitation mechanism to disentangle the id information and activity information. However, their design requires specific network design~(complex network structure) and the additional meta information, which might be less practical. 

\begin{figure*}[t]
    \centering
    \includegraphics[width=\textwidth]{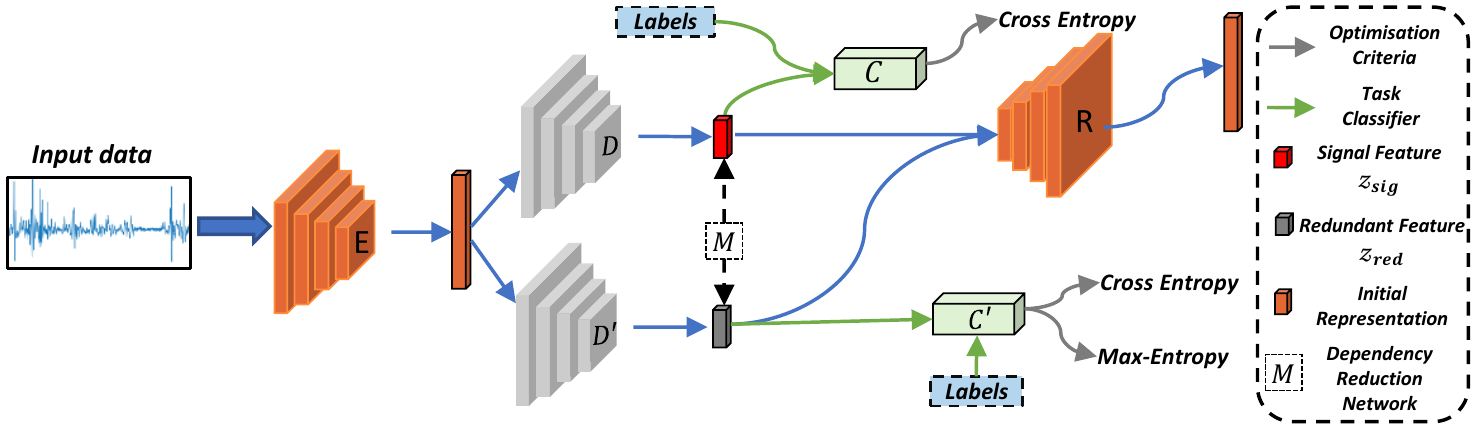}
    \caption{Structure of our proposed BPD framework, where ($E$), ($D$, $D'$), ($C,C'$), ($R$), ($M$)  represent Encoder, Disentanglers, Classifiers, Reconstructor, Dependency Reduction Network, respectively.}
    \label{fig:BPD}
\end{figure*}

\section{Problem Definition}
Recognising human activity with multi-modal data involves multiple devices attached to different parts of the human body. Each device carries multiple sensors~(e.g., 3-axis accelerometer, gyroscope, magnetometer). Following the standard HAR procedure~\cite{bulling2014tutorial}, we divide the multi-variate sensory streams into segments with a fixed-size sliding window~(detailed size information will be listed on experiment setting). Finally, given segment training data $\{\mathbf{x}_i, y_i\}_{i=1}^N$ where $N$ is the training sample number; $\mathbf{x}_i$ and $y_i$ are the $i$th training example and label with $y_i \in [1, K]$ and $K$ is the class number, the purpose of HAR is to learn a function $\mathbb{F}(x,\beta)$ to infer the correct activity label for the given segment data, where $\beta$ represents all the parameters to be learned during the training process.  

\section{Methodology}
For HAR, the collected sensor data often, if not always, includes other redundant features which can be subjects' personal style, gender, age, weight, etc. Such redundant features or factors may cause large intra-activity variability, making it challenging to learn discriminant behaviour patterns, especially when with inadequate data. In order to remove or separate the redundant features from the activity signal, we introduce the Behaviour Pattern Disentanglement (BPD) scheme, based on which the activity signal can be separated from the redundancy features in the latent feature space. 

Our proposed BPD scheme includes two key components as illustrated in Figure~\ref{fig:BPD}: (i) \emph{A Signal \& Redundant feature disentanglement network}, which learns to disentangle the input features into activity signal and redundant features; (ii) \emph{A Dependency Reduction Networks}, which aims to reduce the correlations between the activity signal and redundant features.  These two blocks together with a feature reconstruction module maximises the effects of feature disentanglement while ensuring the minimal information loss. The whole framework can be trained in an end-to-end manner, and the disentanglement networks and the activity signal classifier will be used during inference (i.e., activity classification).

In the following subsections, we will elaborate our design choices on Signal \& Redundant Feature Disentanglement (Section \ref{sec:urd}), the Signal \& Redundant Feature Dependency Reduction (Section \ref{sec:dr}) and the adversarial training and optimisation process (Section \ref{sec:al}).

\subsection{Signal \& Redundant Feature Disentanglement}
\label{sec:urd}
The activity pattern of different users is often associated with users' own personal information, like gender, age and other factors. However, these personalised attributes hinder the representation learning for the later classification. To this end, we propose to utilise a more general and robust feature representation with less irrelevant user information to alleviate the classification difficulty across different users, so as to further improve the generalisation ability of the HAR models. In the following section, we refer to the aforementioned robust feature representation and irrelevant user information as ``activity features'' $\mathbf{z}_{sig}$ and ``redundant features'' $\mathbf{z}_{red}$ respectively. 

We leverage the idea from the generative adversarial network and introduce the concept of \textit{adversarial disentanglement} to remove/disentangle the redundant features from the activity features. We feed the learnt features to disentanglers $D$ and $D'$ to decompose the features representation retrieved from the encoder $E$. The representation extracted from disentangler $D$ will be supervised by the corresponding activity labels in the classifier $C$ to ensure the activity classification ability, while a two-player game happens in the classifier $C'$ and its' adversary disentangler $D'$ so as to generate irrelevant representations. Note that the classifier $C'$ aims to map the representations to the correct activity labels while the $D'$ on the contrary generates the irrelevant representations to fool the classifier $C'$. To guard the representation integrity of the disentangled features from $D$ and $D'$, we add a feature reconstructor $R$ to recover the initial feature representation.

The proposed BPD framework trains the aforementioned encoder ($E$), disentanglers ($D$ and $D'$) and classifiers ($C$ and $C'$) in an alternative way. More precisely, two disentanglers $D$ and $D'$ along with two $K$-way classifiers (i.e., $C$ and $C'$) will be first trained by minimising the cross-entropy loss in Eq.~\eqref{eq:ce}:
\begin{equation}
\begin{aligned}
   \underset{\theta_{E,D,D',C,C'}}{\mathcal{L}_{ce}} = - \frac{1}{N}\sum_{i=1}^{N} \sum^{K}_{k=1}1[y_i=k] \log(C(\mathbf{z}^i_{sig}))
   - \frac{1}{N}\sum_{i=1}^{N}\sum^{K}_{k=1}1[y_i=k]\log(C'(\mathbf{z}^i_{red})) \,.
\label{eq:ce}
\end{aligned}
\end{equation}
where $\mathbf{z}^i_{sig} = D(E(\mathbf{x}_i))$ and $\mathbf{z}^i_{red} = D'(E(\mathbf{x}_i))$.

Then, to produce features with less discriminative information, or equivalently, increasing the uncertainty of classification, we minimise the negative entropy on the feature extracted by disentangler $D'$. The negative entropy function can be written as:
\begin{equation}
   \underset{\theta_{E, D'}}{\mathcal{L}_{ne}} = - \frac{1}{N}\sum_{i=1}^{N} \log C'(\mathbf{z}_{red}^i) \,,
   \label{eq:ne}
\end{equation}
and the parameter of classifiers is fixed when optimising Eq.~\eqref{eq:ne}. 

To ensure the integrity of the feature representation, that is, the disentangled activity and redundant features can be reconstructed back to the initial representation, we forward the features $\mathbf{z}_{sig}$ and $\mathbf{z}_{red}$ that are extracted from disentanglers to the reconstructor $R$ and optimise it simultaneously with the negative entropy minimisation (Eq.~\eqref{eq:ne}). To achieve the effects of reconstruction, we utilise the L2 loss function to constrain the equivalence between reconstructed features and initial features. 
Such constraint can be written as:
\begin{equation}
     \underset{\theta_{D,D',R}}{\mathcal{L}_{recon}} = \frac{1}{N}\sum_{i=1}^{N} \left \|E(\mathbf{x}_i) - R(\mathbf{z}^i_{sig},\mathbf{z}^i_{red})\right \| \,,
\label{eq:recon}
\end{equation}
where $E(x_i)$ is the initial representation. Note, the activity feature $\mathbf{z}_{sig}$ and redundant feature $\mathbf{z}_{red}$ will be fed into the reconstructor with a concatenation operation, which means only one concatenated feature will be forwarded into reconstructor.

\subsection{Signal \& Redundant Feature Dependency Reduction}
\label{sec:dr}

The previous section presents the disentanglement scheme of the activity and redundant features. To ensure less correlated/dependent disentangled features for a good disentanglement learning, we leverage the mutual information---a measure of non-linear dependencies between variables (e.g. learned feature representation)~\cite{kinney2014equitability,sanchez2020learning, ozair2019wasserstein}---to reduce the mutual information between the activity signal $\mathbf{z}_{sig}$ and redundant features $\mathbf{z}_{red}$, so as to reduce the dependence of between them.

Here, the mutual information between the activity features $\mathbf{z}_{sig}$ and redundant features $\mathbf{z}_{red}$ can be defined as: 
\begin{equation}
 I(\mathbf{z}_{sig},\mathbf{z}_{red}) = \int_{\mathbf{z}_{sig}}\int_{\mathbf{z}_{red}} \log \frac{P(\mathbf{z}_{sig},\mathbf{z}_{red})}{P(\mathbf{z}_{sig})P(\mathbf{z}_{red})}d\mathbf{z}_{sig}d\mathbf{z}_{red} \,,
\end{equation}
where $P(\mathbf{z}_{sig},\mathbf{z}_{red})$ is the joint probability density distribution (pdf); $P(\mathbf{z}_{sig})$ and $P(\mathbf{z}_{red})$ are marginal pdfs. $I(\mathbf{z}_{sig},\mathbf{z}_{red})$ measures the dependency between the two features, and  minimising $I(\mathbf{z}_{sig},\mathbf{z}_{red})$ may further push these two features apart in the feature space.

Despite being a pivotal measure across different domains, the mutual information is only tractable for discrete variables, or for a limited family of problems where the probability distributions are unknown~\cite{belghazi2018mutual}. Following \cite{belghazi2018mutual}, we adopt Mutual Information Neural Estimator (MINE) as an unbiased estimation of mutual information on \textit{i.i.d} samples through a neural network $M_{\theta}$ (as shown in Figure \ref{fig:BPD}). Specifically, we leverage the lower-bound calculation from \cite{belghazi2018mutual} to formulate the loss function as follows: 

\begin{equation}
\begin{split}
\underset{\theta_{D,D',M}}{\mathcal{L}_{MINE}} =I(\mathbf{z}_{sig},\mathbf{z}_{red}) = \frac{1}{n} \sum^{n}_{i=1}M(\mathbf{z}^i_{sig},\mathbf{z}^i_{red},\theta) - log(\frac{1}{n} \sum^{n}_{i=1}e^{M(\mathbf{z}^i_{sig},\mathbf{\hat{z}}^i_{red},\theta)}) \,,
\label{eq:mine}
\end{split}
\end{equation}
where $\{\mathbf{z}^i_{sig},\mathbf{z}^i_{red}\}_{i=1}^n$ are $n$ pairs sampled from the joint distribution $P(\mathbf{z}_{sig},\mathbf{z}_{red})$; $\mathbf{\hat{z}}^i_{red}$ is sampled from the marginal distribution $P(\mathbf{z}_{red})$, and $M(\mathbf{z}^i_{sig},\mathbf{z}^i_{red},\theta)$ is a neural network parameterised by $\theta$ to estimate the mutual information between two distributions.
Model parameters of the disentanglers ($D$ and $D'$) as well as the dependency reduction network $M$ will be updated by minimising Eq.(\ref{eq:mine}).

\begin{algorithm}[t]
\SetAlgoLined
\SetKwInOut{Input}{Input}
\Input{ Training data $\{\mathbf{x}_i,y_i\}_{i=1}^N$, Encoder $E$, Disentanglers ($D$ and $D'$); Classifiers ($C$ and $C'$); Dependency Reduction Network $M$, and Reconstructor $R$;}
\KwResult{Trained Encoder $E$, trained disentangler $D$ and trained classifier $C$}
 \textbf{Initialisation}\;
 \For{ j=1 : maxEpoch}{
  \uIf{not converged}{
  Sampling training mini-batch data from $\{\mathbf{x}_i,y_i\}_{i=1}^N$\;
  \textbf{Signal$\&$Noise Disentanglement:} \\
  Updating $E,D,D',C, C'$ by minimising Eq. (\ref{eq:ce})\;
  Updating $E, D'$ by minimising Eq. (\ref{eq:ne})\;
  \textbf{Signal$\&$Noise Dependency Reduction:}\\
  Updating $D, D', M$ by minimising Eq. (\ref{eq:mine}) \;
  \textbf{Reconstruction:}\\
  Updating $D,D', R$ by minimising Eq. (\ref{eq:recon}) \;
  $j=j+1$ \;
  }
  \Else{break;}
 }
 \Return Trained $E$; trained $C$; trained $D$
 
\caption{Training Strategy of the BPD framework}
\label{al:BPD}
\end{algorithm}

\subsection{Algorithm \& Implementation}
\label{sec:al}
For our BPD structure, we implement the components as follows: 1) Encoder~($E$): CNN\cite{2015MCCNN} or DeepConvLSTM\cite{2016DeepConvLSTM} with the same network structure as original works; 2) Disentanglers ($D$ and $D'$): Single fully-connected layer with a batch normalisation layer; 3) Reconstructor~($R$): single fully-connected layer. 4) Dependency Reduction Network ($M$): two fully-connected layers. 5) Classifiers~($C$ and $C'$): two fully-connected layers and a dropout function. To ensure the learned consistent representation space can lead to accurate activity classification, we jointly minimise the activity classification error~(Classifier $C$) and conduct adversarial training. 

The training strategy of our BPD framework is detailed in Algorithm \ref{al:BPD}. More precisely, on the training stage, given data from training users, with selected Encoder $E$~(e.g., CNN, ConvLSTM), the BPD framework is optimised in an end-to-end manner using Algorithm \ref{al:BPD}. At the inference stage, given trained components $E$~(CNN or ConvLSTM), $C$~(Classifier), $D$~(Disentangler) from Algorithm \ref{al:BPD}, for any query data $\mathbf{x}$ from any unseen user, classification can be performed by inputting the disentangled activity features to the classifier $C$.
Specifically, the label is assigned to $\hat{y}$ such that:
\begin{equation}
\hat{y} = \arg\max_{[1,K]}C(\mathbf{z}_{sig}),\quad \text{where} \quad \mathbf{z}_{sig} = D(E(\mathbf{x})).
\label{eq:inference}
\end{equation}

\section{Experiment}
\subsection{Datasets}
To evaluate the effectiveness of our BPD framework, we perform it on four public datasets: PAMAP2~\citep{PAMAP2}, MHEALTH~\citep{MHEALTH}, DSADS~\citep{UCIDSADS} and GOTOV~\citep{GOTOV}.

\begin{table}[h]
\caption{Description of the four public HAR datasets used in our study; \#Dim represents the dimension of the input data.}
\label{tb:dataset}
\begin{tabular}{@{}ccccccc@{}}
\toprule
Dataset & \#Subject & \#Activity & Frequency & \#Sample& \#Dim & Wearing Position                                      \\ \midrule
PAMAP2  & 8       & 12       & 100Hz     & 2.84M & 52 & Wrist,Chest,Ankle                             \\
MHEALTH & 10      & 12       & 50Hz      & 0.34M  & 23 & Chest,Ankle,Arm                               \\
DSADS   & 8       & 19       & 25Hz      & 1.14M & 45 & Tarso, Right/Left Arm, Right/Left Leg \\ 
GOTOV   & 35      &16        & 83HZ      & 5.9M & 3 & Wrist\\\bottomrule
\end{tabular}
\end{table}

\begin{figure}[b]
    \centering
    \includegraphics[width=\textwidth]{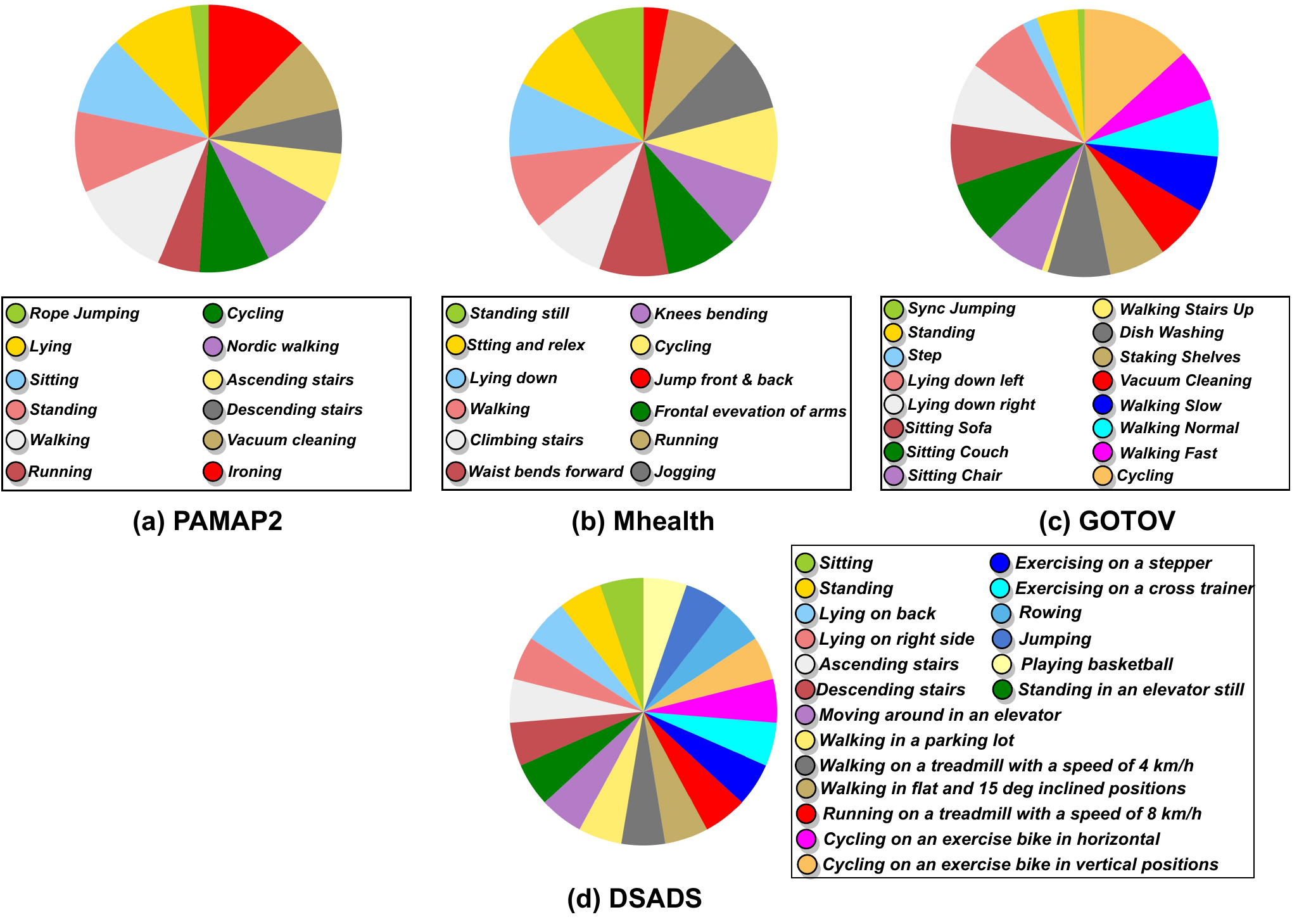}
    \caption{Activity distribution of the four datasets in our study (best viewed in colour)}
    \label{fig:data_dis}
\end{figure}

\textit{Physical Activity Monitoring} (PAMAP2)~\citep{PAMAP2} dataset includes data recorded from 9 subjects performing 18 different activities, such as vacuum cleaning, ironing, rope jumping, etc. The data were collected with three IMUs placed on the subject's chest, dominant wrist, and dominant ankle, respectively. In our study, 12 activities were selected (as shown in Fig. \ref{fig:data_dis}), and all the IMU data channels (i.e., 52 dimensions) from 8 subjects were used. 

\textit{Mobile Health} (MHEALTH)~\citep{MHEALTH} dataset contains body motion and vital signs recording for 10 subjects of diverse profiles while performing 12 activities in an out-of-lab environment with no constraints. The total dimension of the input data is 23, which include the data recorded by inertial measurement units~(IMUs) that are placed on the subject's chest, right wrist and left ankle. The IMUs collect a 3-axis acceleration, a 3-axis gyroscope and a 3-axis magnetic field of motion, respectively. Also, the IMUs positioned on the chest provides 2-lead ECG measurement, which can be used for basic heart monitoring.

\textit{Daily and Sports Activities Data Set} (DSADS)~\citep{UCIDSADS} dataset contains motion sensor data of 19 daily and sports activities performed by 8 subjects. Each activity was performed for 5 minutes in their style without constraints. 5 IMUs were positioned on the torso, right arm, left arm, right leg and left leg with 9 sensors on each unit~(3-axis accelerometers, 3-axis gyroscopes, and 3-axis magnetometers) which produces 45-dimensional sensor data.

\textit{Growing Old Together Validation} (GOTOV)~\citep{GOTOV} dataset contains 16 daily activities sensor data of 35 elder participants (21 male and 14 female). The inertial measurement units were placed on the subject's ankle, chest, and wrist to collect 3-axis acceleration data. 

For DSADS dataset, we used all subjects' data. For PAMAP2, we removed 6 activities (i.e., Watching TV, Computer work, Car driving, Folding laundry, House Cleaning, and Playing soccer), as they were only performed by one subject, which was also removed in our study. For MHEALTH dataset, we used all subjects' data. 
For the GOTOV dataset, since there are some missing channels in the sensors attached to ankle/chest, only wrist-worn accelerometer data (i.e., with 3 dimensions) were used in our study.
Details of these used datasets can be found in Table~\ref{tb:dataset}.

Figure~\ref{fig:data_dis} presents the activity distribution for the four datasets. In terms of activity classes, DSADS dataset is more balanced than the other three. Since the unbalanced class distribution might affect the performance of the algorithms, we further report the class-wise performance.

\subsection{Experimental Settings}
\paragraph{Data Pre-processing}
In our study, we divided the raw sensory data streams into small data segments with a fixed-sized sliding window (168 samples) with an overlap of $50\%$. Since the sampling rates for the four datasets are different, it yields window lengths of 1.68, 3.36, 6.72, 2.02 seconds for the PAMAP2, MHEALTH, DSADS, and GOTOV datasets, respectively. These windows/segments can be fed into the network directly without any hand-crafted feature engineering or transformation. 

\paragraph{Baseline Models}We compared our proposed BPD framework with the closely related baselines. CNN\cite{2015MCCNN} and DeepConvLSTM\cite{2016DeepConvLSTM} were the state-of-the-art feature learning approaches for human activity recognition; beta-VAE\cite{higgins2016beta} was a conventional disentangle learning framework in the computer vision field and GILE ~\cite{2021GILE} was a recent method aiming at disentangling the identity information from raw data streams, yet subject identity label is required by GILE as an extra information. For all baseline methods, we used the released code if available, and reproduced the unavailable methods using Pytorch\cite{pytorch}.

\paragraph{Training Setting} Our network parameters were initialised by Xavier Normal~\cite{glorot2010understanding} and optimised by Adam Optimiser~\cite{kingma2014adam} with a learning rate of 0.0001 for four datasets. Due to the computational limitation, we set the training batch size to 64 with $300$ maximum training epoch. For the latent representation dimension~(i.e., feature dimension of activity code  $\mathbf{z}_{sig}$/redundant code $\mathbf{z}_{red}$), we empirically set $592$ to encoder CNN and $32$ to DeepConvLSTM respectively. All algorithms were implemented by Pytorch and running on NVIDIA RTX 3090 GPU.

\paragraph{Evaluation Protocol} Initially, we used the Leave-One-Subject-Out Cross Validation~(LOSO-CV) strategy for each dataset to evaluate the models' performance. For each dataset, both the overall performance and the subject-level performance were reported. Moreover, we conducted the ablation study on the large GOTOV dataset (35 subjects in total) using the hold-out validation, where the training set included 28 subjects while the test set included the other 7 subjects. 

\paragraph{Evaluation Metric} 

To measure the performance of our proposed BPD framework, we used the mean F1 score as the evaluation metric, which is widely used in the human activity recognition literature~\cite{2017ensembleslstm,2016DeepConvLSTM}. Moreover, we also reported the class-wise F1-score for the four datasets to investigate the effect of redundant information on the class level~(i.e., to see which class benefits more when removing the irrelevant information) for a better understanding of our BPD framework.

\subsection{Result on Four public HAR Datasets}
We evaluated our BPD framework on the four public HAR datasets. 
In Table~\ref{tb:pamap2-loso},~\ref{tb:mhealth-loso},~\ref{tb:dsads-loso}, the mean F1-scores of the baseline models as well as our BPD framework~(based on two encoders, i.e., CNN and DeepConvLSTM) were reported in both subject-level and overall average for datasets PAMAP2, MHEALTH, and DSADS. 
We can see the superior performance improvements in all settings brought by the proposed BPD framework, irrespective of the encoders. 
We also compared two disentanglement learning baselines, namely, beta-VAE\cite{higgins2016beta} and GILE ~\cite{2021GILE}.
From these tables, we can see both methods yield lower performance even than the baselines without disentanglement. Since beta-VAE was an approach borrowed from the computer vision field, it may not generalise well to HAR tasks.
On the other hand, the low performance of the GILE might be due to the enforcement of mapping features into subject-identifiable-level
which may be unnecessary and hard to train. 
We grid-searched the hyper-parameters of GILE for the best results, yet the results were less promising when compared with our approach. 
Moreover, GILE requires human identity labels, which can be less flexible than ours.

\begin{table}[t]
\centering
\caption{Mean F1-score for each subject on the PAMAP2 dataset (in leave-one-subject-out CV setting)}
\label{tb:pamap2-loso}
\begin{tabular}{@{}c|c|cccccc@{}}
\toprule
Dataset                 & Subject & CNN   & DeepConvLSTM & beta-VAE &\hspace{0.5cm}GILE\hspace{0.5cm} & \begin{tabular}[c]{@{}c@{}}BPD\\ (CNN)\end{tabular} & \begin{tabular}[c]{@{}c@{}}BPD\\ (DeepConvLSTM)\end{tabular} \\ \midrule
\multirow{9}{*}{PAMAP2} & 1       & 0.6539 & 0.6340       & 0.5970   & 0.6032 & 0.6826 & \textbf{0.6915} \\
                        & 2       & 0.7563 & 0.7363       & 0.7084   & 0.7181 & \textbf{0.8719} & 0.8381          \\
                        & 3       & 0.8099 & 0.7154       & 0.5713   & 0.6950 & \textbf{0.8262} & 0.8117          \\
                        & 4       & 0.8044 & 0.8183       & 0.7115   & 0.7542 & \textbf{0.8307} & 0.8244          \\
                        & 5       & 0.8886 & 0.8588       & 0.7391   & 0.8181 & \textbf{0.8900} & 0.8675          \\
                        & 6       & 0.8791 & 0.7924       & 0.7175   & 0.7723 & \textbf{0.8827} & 0.8281          \\
                        & 7       & 0.9243 & 0.9100       & 0.8376   & 0.8932 & \textbf{0.9311} & 0.9101 \\
                        & 8       & 0.3952 & 0.4495       & 0.3814   & 0.3521 & 0.4011          & \textbf{0.4921} \\\cmidrule(l){2-8}
                        & Avg.    & 0.7640 & 0.7393       & 0.6580   & 0.7008 & \textbf{0.7895} & 0.7829 
                                              \\ \cmidrule(l){1-8} 
\end{tabular}
\end{table}

\begin{table}[]
\centering
\caption{Mean F1-score for each subject on the Mhealth dataset (in leave-one-subject-out CV setting)}
\label{tb:mhealth-loso}
\begin{tabular}{@{}c|c|cccccc@{}}
\toprule
Dataset                   & Subject & CNN   & DeepConvLSTM & beta-VAE & \hspace{0.5cm}GILE\hspace{0.5cm} & \begin{tabular}[c]{@{}c@{}}BPD\\ (CNN)\end{tabular} & \begin{tabular}[c]{@{}c@{}}BPD\\ (DeepConvLSTM)\end{tabular} \\ \midrule
\multirow{11}{*}{MHEALTH} & 1       & 0.9514 & 0.9074       & 0.7866   & 0.8452 & \textbf{0.9575} & 0.9554          \\
                          & 2       & 0.8530 & 0.8760       & 0.8195   & 0.8135 & \textbf{0.9348} & 0.9085          \\
                          & 3       & 0.8441 & 0.8688       & 0.7091   & 0.8322 & 0.8659          & \textbf{0.8711} \\
                          & 4       & 0.9351 & 0.9112       & 0.8101   & 0.8810 & 0.9510          & \textbf{0.9573} \\
                          & 5       & 0.8781 & 0.8583       & 0.7752   & 0.8150 & \textbf{0.9904} & 0.9804          \\
                          & 6       & 0.9849 & 0.9241       & 0.8753   & 0.8832 & \textbf{0.9934} & 0.9766          \\
                          & 7       & 0.9793 & 0.9735       & 0.6985   & 0.8923 & \textbf{0.9965} & 0.9760          \\
                          & 8       & 0.9685 & 0.9566       & 0.8706   & 0.8842 & \textbf{0.9848} & 0.9775          \\
                          & 9       & 0.9894 & 0.9812       & 0.9113   & 0.9237 & \textbf{0.9913} & 0.9869          \\
                          & 10      & 0.9829 & 0.9383       & 0.8737   & 0.9224 & \textbf{0.9943} & 0.9865          \\\cmidrule(l){2-8}
                          & Avg.    & 0.9367 & 0.9195       & 0.8130   & 0.8693 & \textbf{0.9660} & 0.9576 
                                                      \\ \cmidrule(l){1-8} 
\end{tabular}
\end{table}

\begin{table}[h]
\centering
\caption{Mean F1-score for each subject on the DSADS dataset (in leave-one-subject-out CV setting)}
\label{tb:dsads-loso}
\begin{tabular}{@{}c|c|cccccc@{}}
\toprule
Dataset                & Subject & CNN   & DeepConvLSTM & beta-VAE &\hspace{0.5cm} GILE  \hspace{0.5cm} & \begin{tabular}[c]{@{}c@{}}BPD\\ (CNN)\end{tabular} & \begin{tabular}[c]{@{}c@{}}BPD\\ (DeepConvLSTM)\end{tabular} \\ \midrule
\multirow{9}{*}{DSADS} & 1         & 0.7354 & 0.7316        & 0.6135    & 0.7467      & \textbf{0.7599}                                      & 0.7267                                                        \\
                       & 2         & 0.7661 & 0.7804        & 0.6545    & 0.7746 & \textbf{0.8919}                                               & 0.8850                                               \\
                       & 3         & 0.7468 & 0.8327        & 0.5858    & 0.7305 & 0.8755                                     & \textbf{0.8947}                                                        \\
                       & 4         & 0.6734 & 0.6551        & 0.5012    & 0.6579       & 0.6626                                               & \textbf{0.6756}                                               \\
                       & 5         & 0.6472 & 0.6504        & 0.4915    & 0.7228 & \textbf{0.7894}                                      & 0.7661                                                        \\
                       & 6         & 0.8023 & 0.9027        & 0.5225    & 0.8185 & \textbf{0.9551}                                      & 0.9302                                                        \\
                       & 7         & 0.7242 & 0.7341        & 0.5724    & 0.6363 & \textbf{0.8615}                                      & 0.8537                                                        \\
                       & 8         & 0.6139 & 0.6255        & 0.5584    &  0.5875   & \textbf{0.7498}                                               & 0.7247                                               \\\cmidrule(l){2-8} 
                       & Avg.      & 0.7136 & 0.7390        & 0.5625    & 0.7094 & \textbf{0.8182}                                               & 0.8070                                               \\ \cmidrule(l){1-8} 
\end{tabular}
\end{table}

In Table~\ref{tb:pamap2-loso}, we also noticed the low results from subject 8. 
Although the BPD scheme can refine the activity feature and improve the results substantially~(about $2-4\%$), they are still far from satisfactory. One major reason can be the limited number of subjects for training. In the LOSO-CV setting, only 7 subjects were used for training, and the trained model may not generalise well to unseen subjects that are very different from the (small) population.

On the MHEALTH and DSADS datasets, although with different activity types, from Table ~\ref{tb:mhealth-loso}, and Table~\ref{tb:dsads-loso}, we can see more significant results: 1) BPD can boost the performance much further, irrespective of the encoder. 2) when compared with other disentanglement learning baselines (beta-VAE, GILE), our BPD yields much higher results.

Compared with PAMAP2, MHEALTH, and DSADS datasets, GOTOV is a much larger dataset with 35 subjects, based on which we conducted LOSO-CV and reported the corresponding results in Table \ref{appdix:result} on appendix \ref{appdix_gotov}. We can observe that the results can benefit from our BPD scheme, with more significant improvement on the CNN encoder than the DeepConvLSTM encoder. Specifically, as shown in Table \ref{appdix:result}, BPD(CNN) and BPD(DeepConvLSTM) can yield about $3.08\%$, and $2\%$ performance gain (in terms of overall average), respectively. 

\subsection{Class-wise Analysis}
Previous experimental results present the performance in both overall and subject-level. To get more insight into our BPD scheme, it is also crucial to show the class-wise or activity-wise results to see how BPD can disentangle redundancy from different activities. 

\begin{figure}[h]
    \centering
    \begin{subfigure}[b]{0.42\textwidth}
        \includegraphics[width=\textwidth]{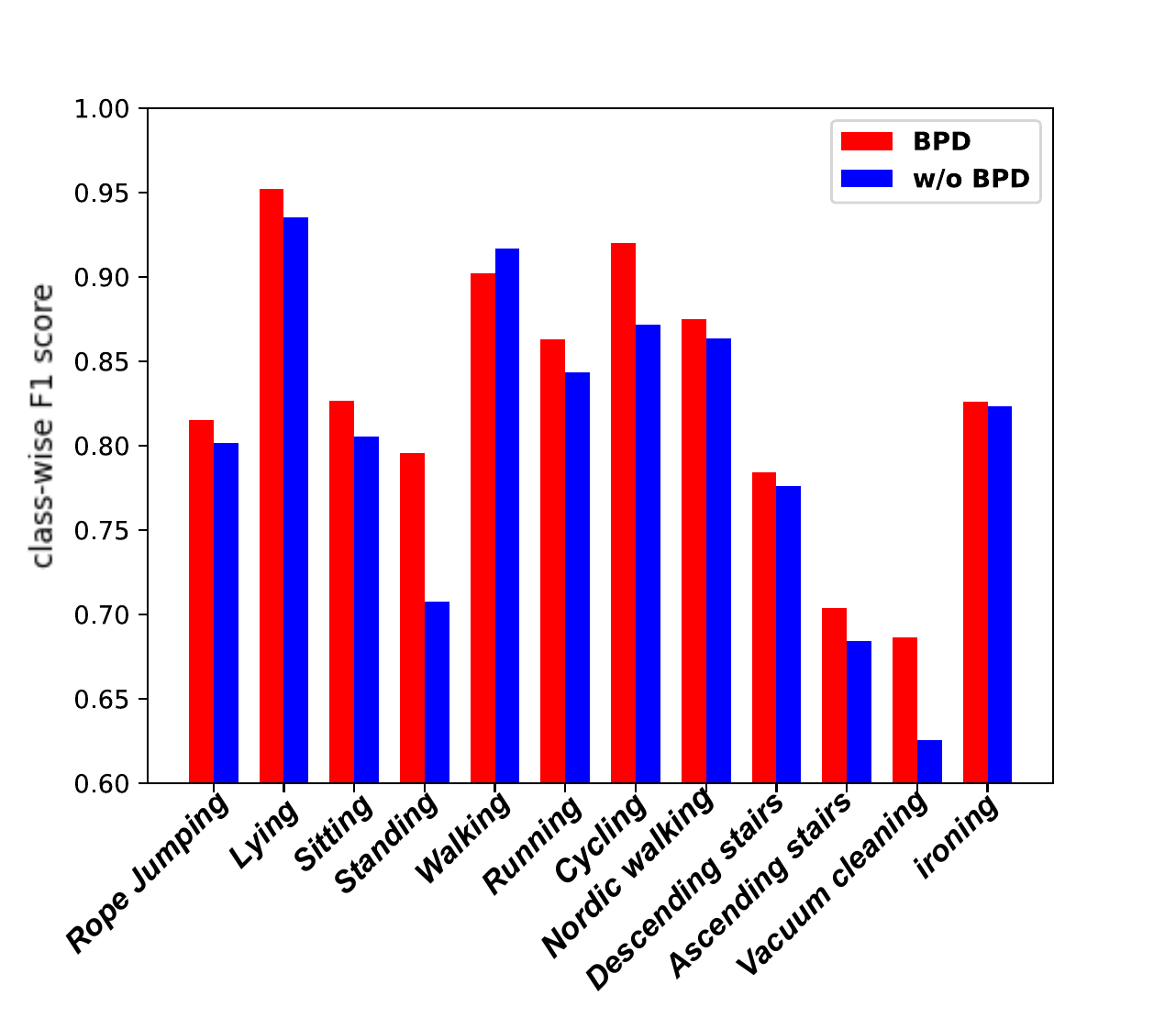}
        \caption{PAMAP2}
        \label{fig:pc_pamap2}
    \end{subfigure}
    \begin{subfigure}[b]{0.41\textwidth}
        \includegraphics[width=\textwidth]{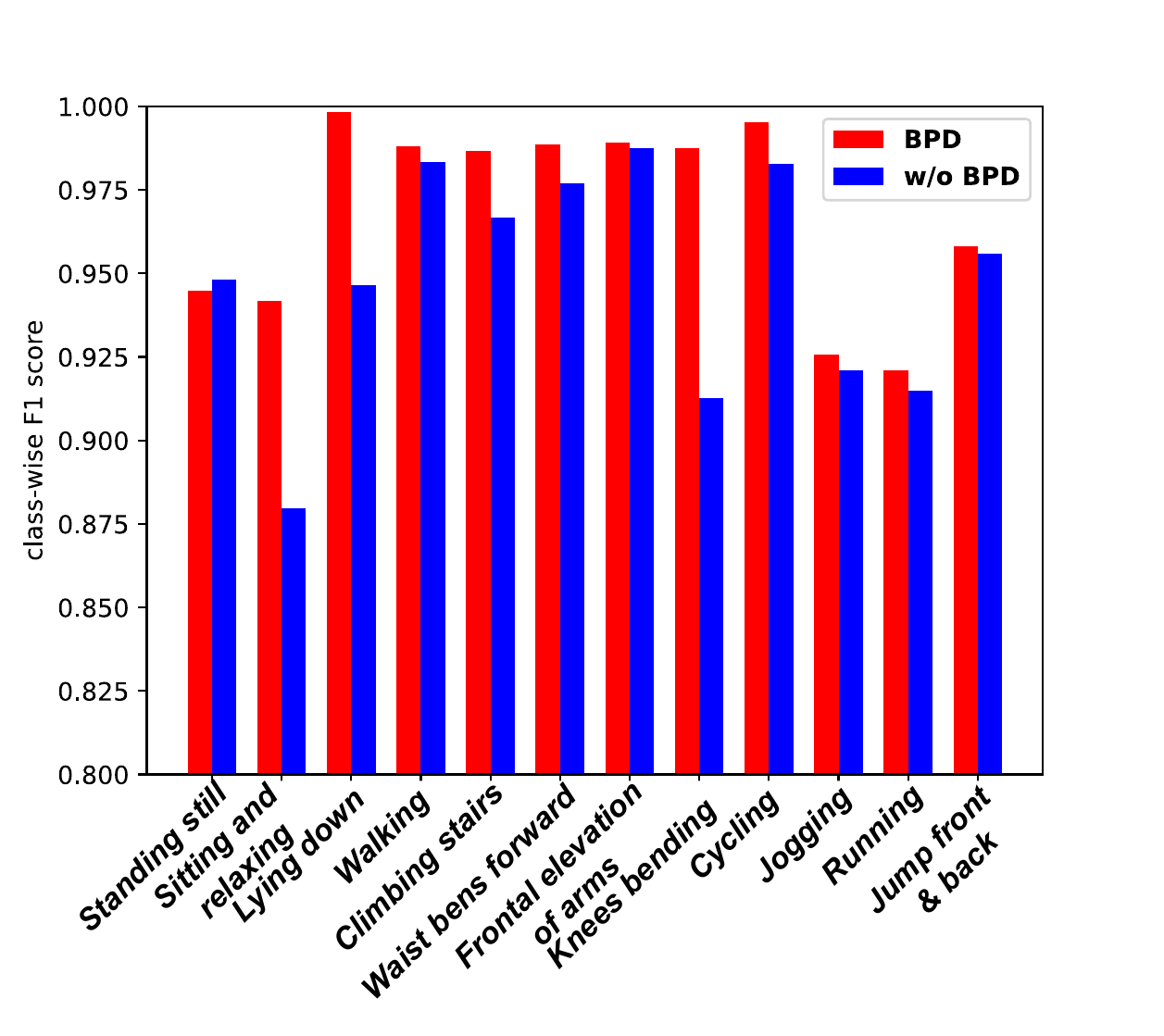}
        \caption{MHEALTH}
        \label{fig:pc_mhealth}
    \end{subfigure}
       \medskip
    \begin{subfigure}[b]{0.42\textwidth}
    \includegraphics[width=\textwidth]{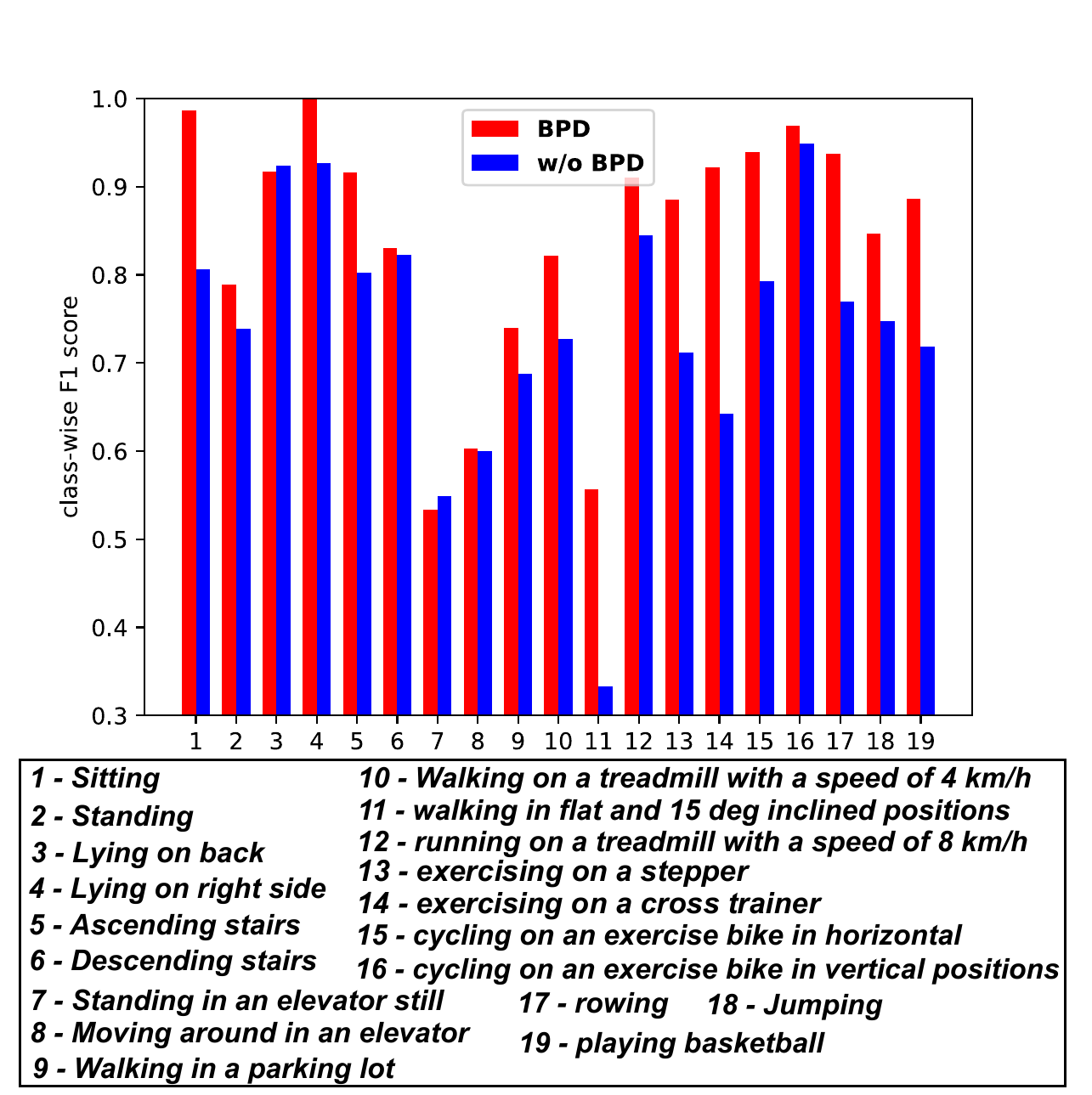}
        \caption{DSADS}
        \label{fig:pc_dsads}
    \end{subfigure}
    \begin{subfigure}[b]{0.42\textwidth}
    \includegraphics[width=\textwidth]{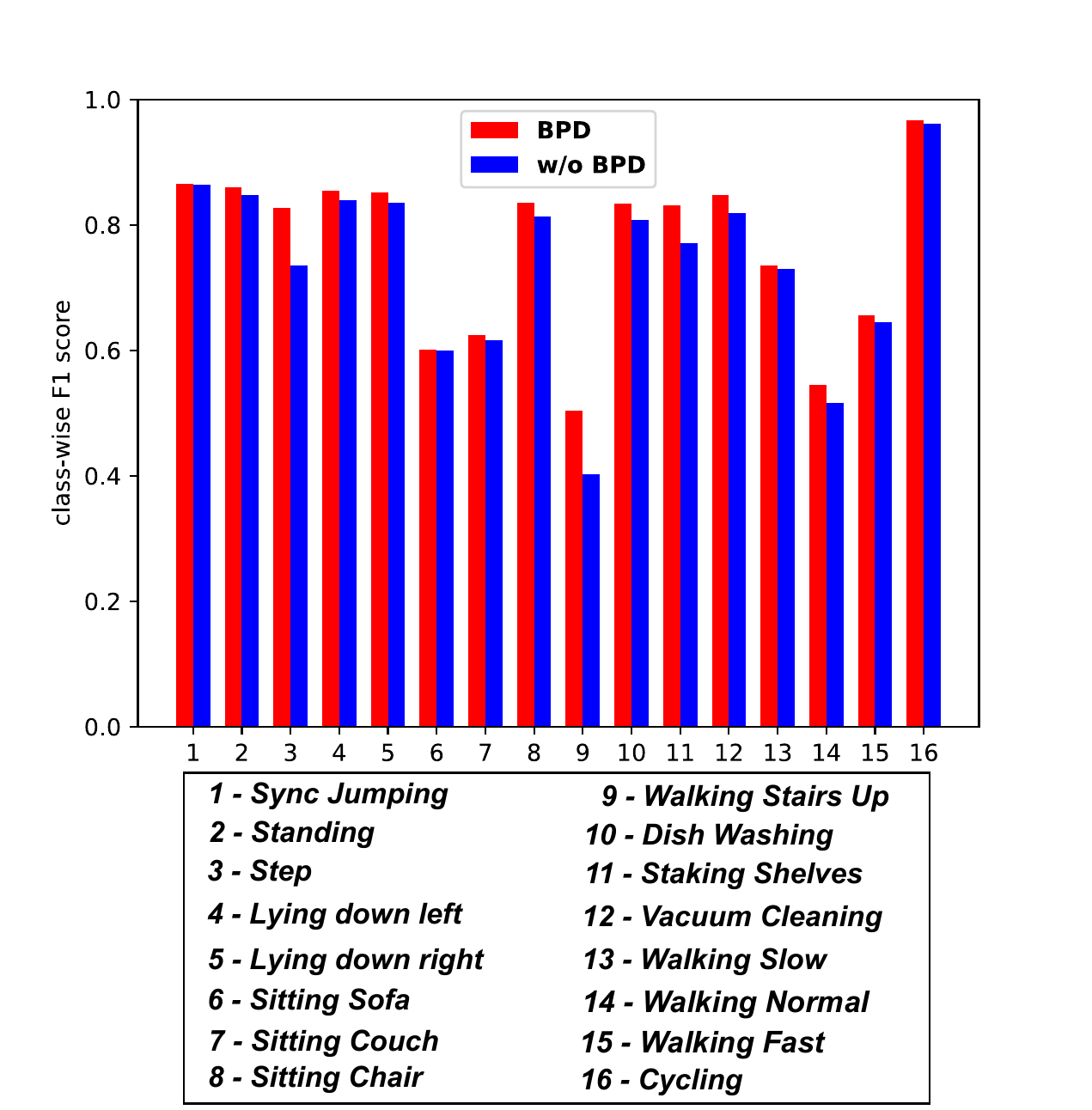}
        \caption{GOTOV}
        \label{fig:pc_gotov}
    \end{subfigure}
    \caption{Class-wise F1-score of CNN with and w/o the proposed BPD scheme on the four datasets}
    \label{fig:pc_f1}
\end{figure}

We conducted the experiments using CNN and BPD(CNN) on the four datasets, and reported the class-wise results in Figure \ref{fig:pc_f1}. 
We can see the general performance gains (by using BPD), yet they vary at the activity level. 
For the PAMAP2 dataset, substantial improvements are from the following activities: ``Vacuum cleaning'', ``Standing'', ``Cycling'', ``Ascending stairs''. 
It is quite interesting to see that in the MHEALTH, DSADS and GOTOV datasets (Figure \ref{fig:pc_mhealth},\ref{fig:pc_dsads}, \ref{fig:pc_gotov}), the ``Ascending stairs'' activity~(or analogously, ``Climbing Stairs'' on MHEALTH dataset and ``Walking Stairs Up'' on GOTOV dataset) were also the activities that benefit significantly from the BPD scheme, indicating these activities may be easily affected by personal/environmental factors. 
Similarly, some strenuous activities such as ``Cycling'', `Vacuum cleaning''  , ``Knees bending'', ``Exercising'', and `` Rowing'' are more likely to be affected by physical factors such as vital capacity so that might cause large variance for different subjects. 
On the contrary, the activities with less energy consumption~(e.g., ``Lying'' activities) will benefit less from the BPD framework since the patterns for those activities tend to be less personal. It is interesting to see that the ``Standing'' activity on PAMAP2, and ``Sitting'' activity on MHEALTH and DSADS dataset gain large improvement. A possible explanation for that is these activities may be affected by the personal habit~(i.e., sitting/standing posture). 

\subsection{Ablation Study}
To study the effectiveness of the major components in the BPD framework, we conducted ablation studies.
Due to the computational limitation, we used the GOTOV dataset with a hold out setting. The training group contains 28 subjects (17 males and 11 females) while the testing group contains 7 subjects (4 males and 3 females). 

\begin{figure}[h]
    \centering
    \includegraphics[width=0.6\textwidth]{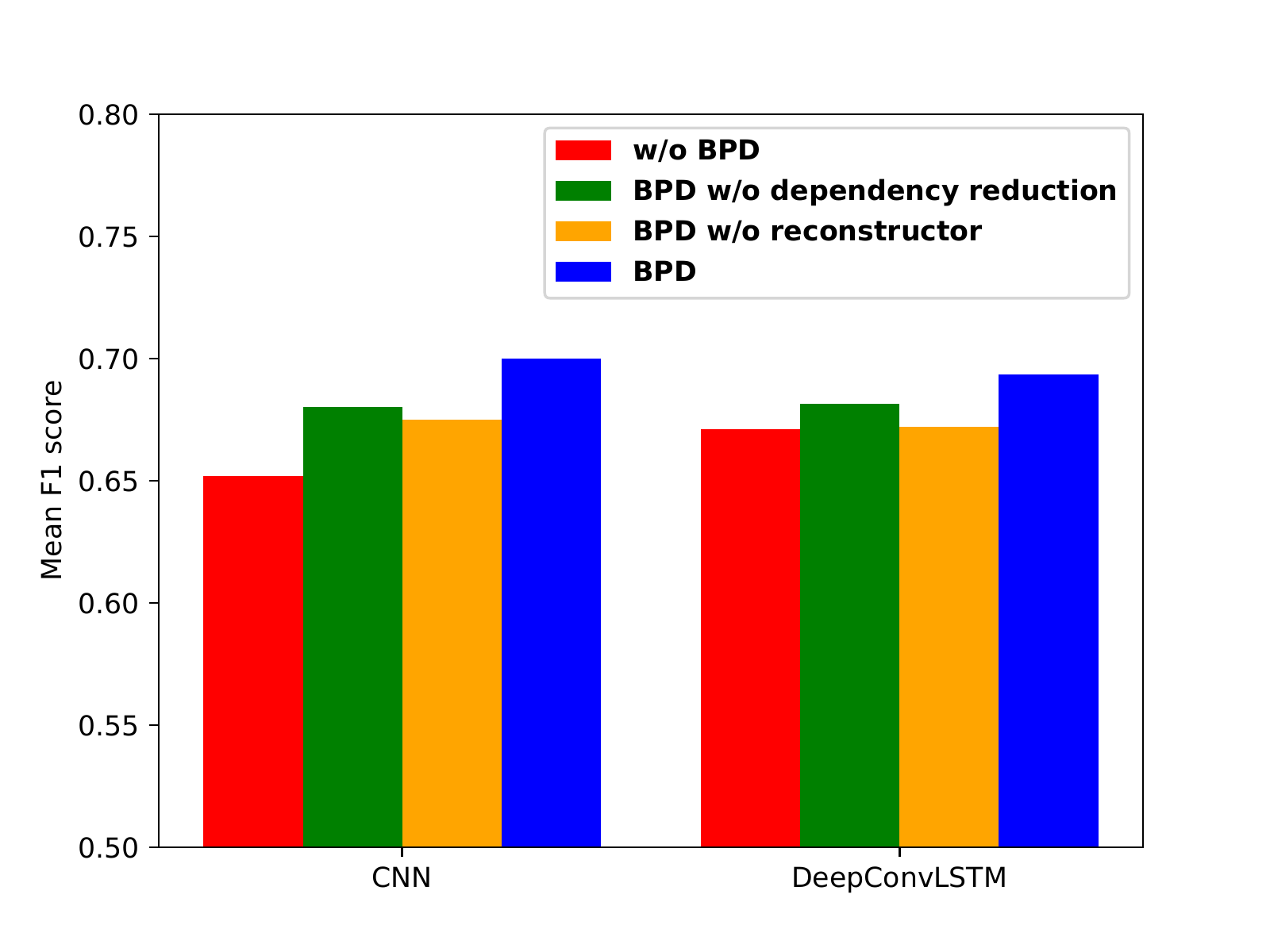}
    \caption{Mean F1-score of the ablation study on GOTOV datasets}
    \label{fig:gotov_ab}
\end{figure}

With encoders (CNN\cite{2015MCCNN} and DeepConvLSTM\cite{2016DeepConvLSTM}), we studied the baseline (i.e., w/o BPD), the signal $\&$ redundancy disentanglement component (i.e., BPD w/o dependency reduction), and the proposed full BPD.
In addition, we also studied the contribution of reconstructor (which can keep the integrity of the feature representation) by removing it from BPD~(i.e., BPD w/o reconstructor).

Figure \ref{fig:gotov_ab} reports the detailed results of the ablation studies, and we can see that each component in the BPD framework contributes positively to the final results, and the encoder CNN benefits more from the BPD framework than DeepConvLSTM.
For encoder CNN, the application of the signal $\&$ redundancy disentanglement component  (i.e., BPD w/o dependency reduction) can achieve about $3\%$ performance improvement, in contrast to only $1\%$ for DeepConvLSTM.
For both encoders, performing the dependency reduction mechanism can further push signal $\&$ redundancy features apart, with further performance gains (i.e., BPD in Fig. \ref{fig:gotov_ab} ). 
We can also observe that for both encoders, the performance of the BPD drop substantially (about 2\% in mean F1 score) without using reconstructor (i.e., BPD w/o reconstructor in Fig.\ref{fig:gotov_ab}), indicating the 
importance of keeping the integrity of the feature representation. 

\subsection{Disentanglement Analysis}
To further verify the effectiveness of our BPD framework--(whether the intra-class variability is reduced), we applied t-SNE to generate visualisation on latent features for GOTOV dataset.

\begin{figure}[h]
    \begin{subfigure}[b]{\textwidth}
        \includegraphics[width=\textwidth]{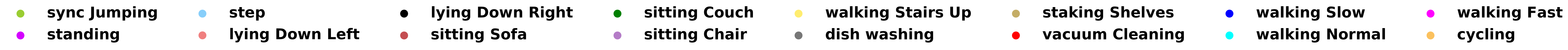}
        \label{fig:leg}
    \end{subfigure}
    \smallskip
    \begin{subfigure}[b]{0.45\textwidth}
        \includegraphics[width=\textwidth]{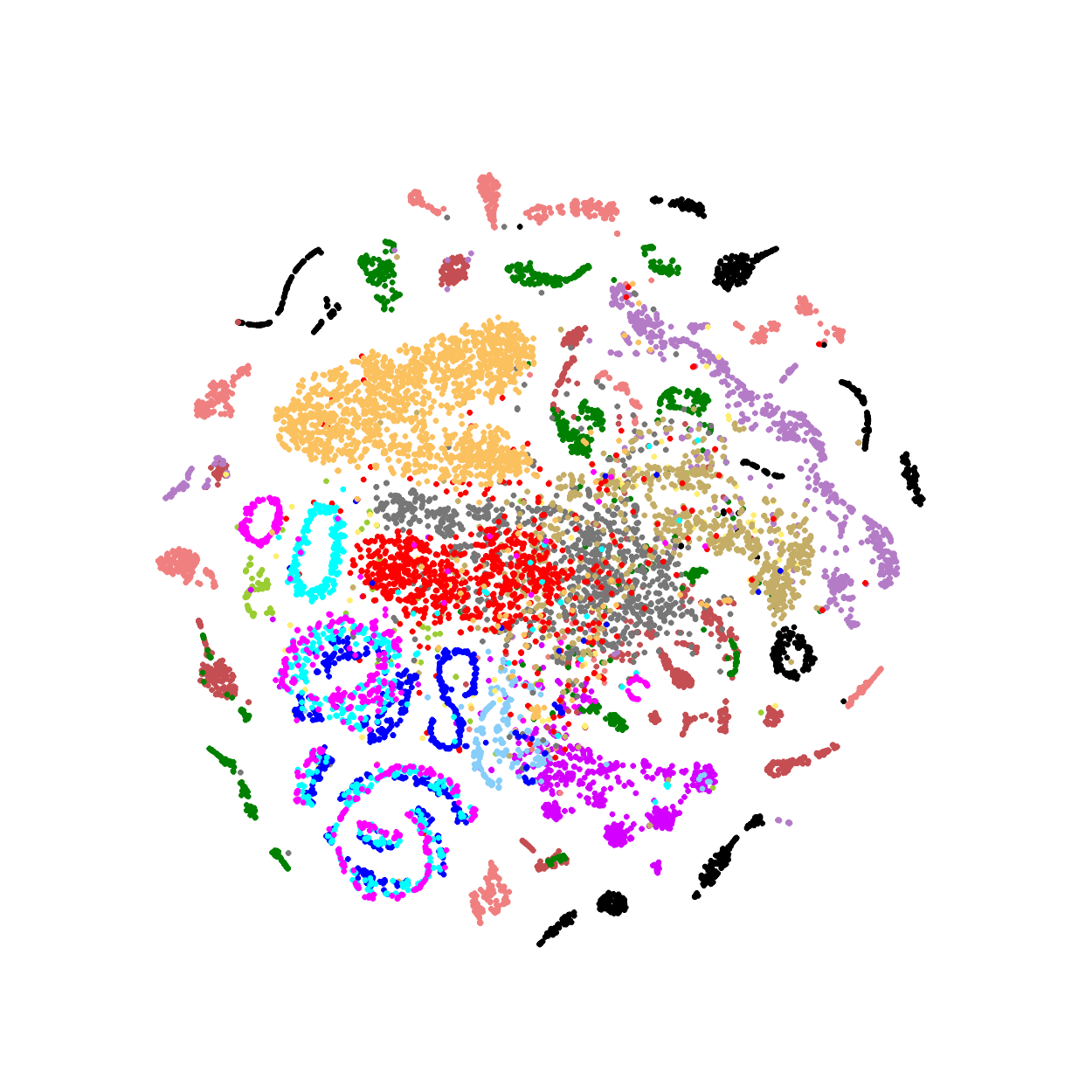}
        \caption{w/o BPD(CNN)}
        \label{fig:cnn}
    \end{subfigure}
    \centering
    \begin{subfigure}[b]{0.45\textwidth}
        \includegraphics[width=\textwidth]{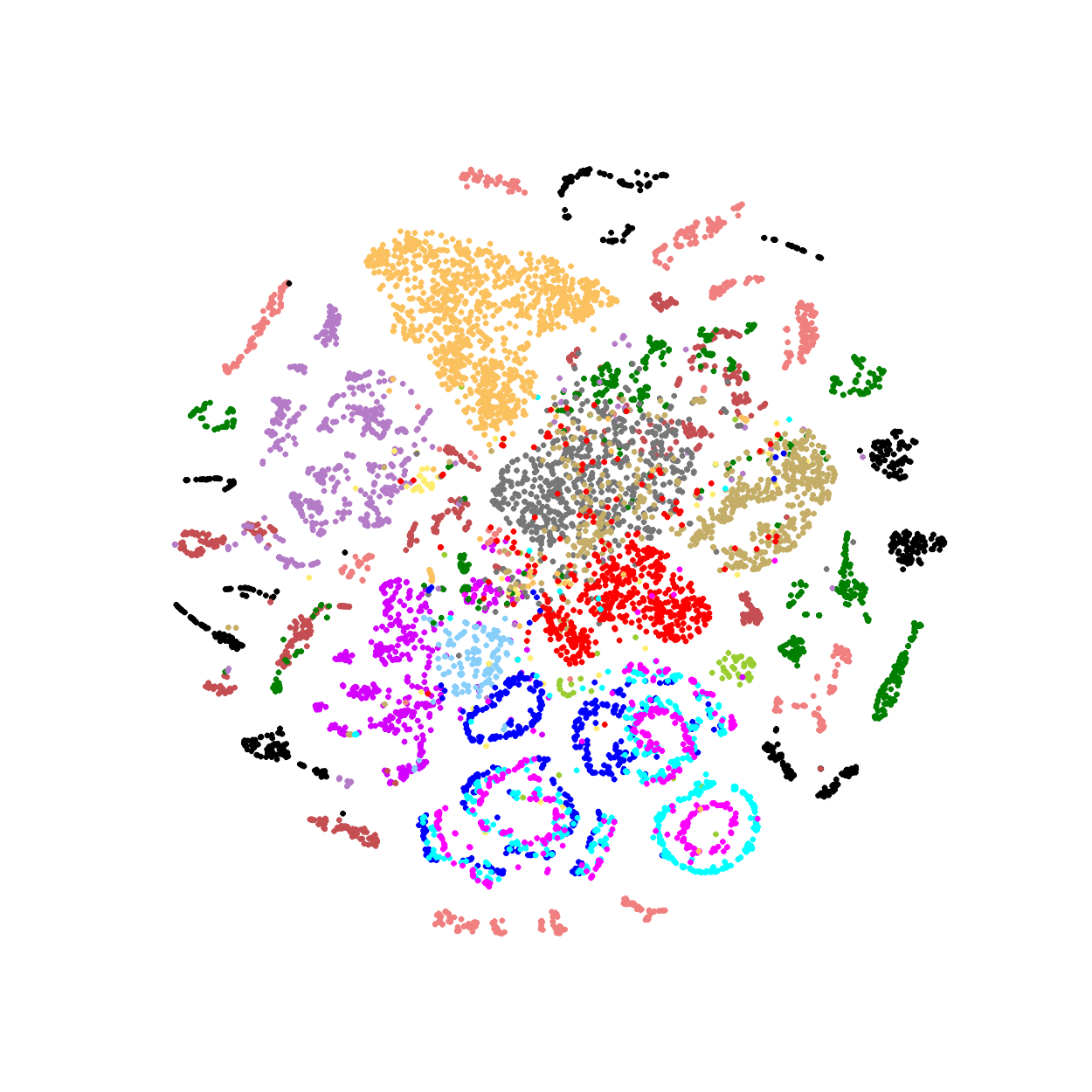}
        \caption{BPD(CNN) Activity Features}
        \label{fig:bpd_ai}
    \end{subfigure}
    \medskip
    \begin{subfigure}[b]{0.45\textwidth}
        \includegraphics[width=\textwidth]{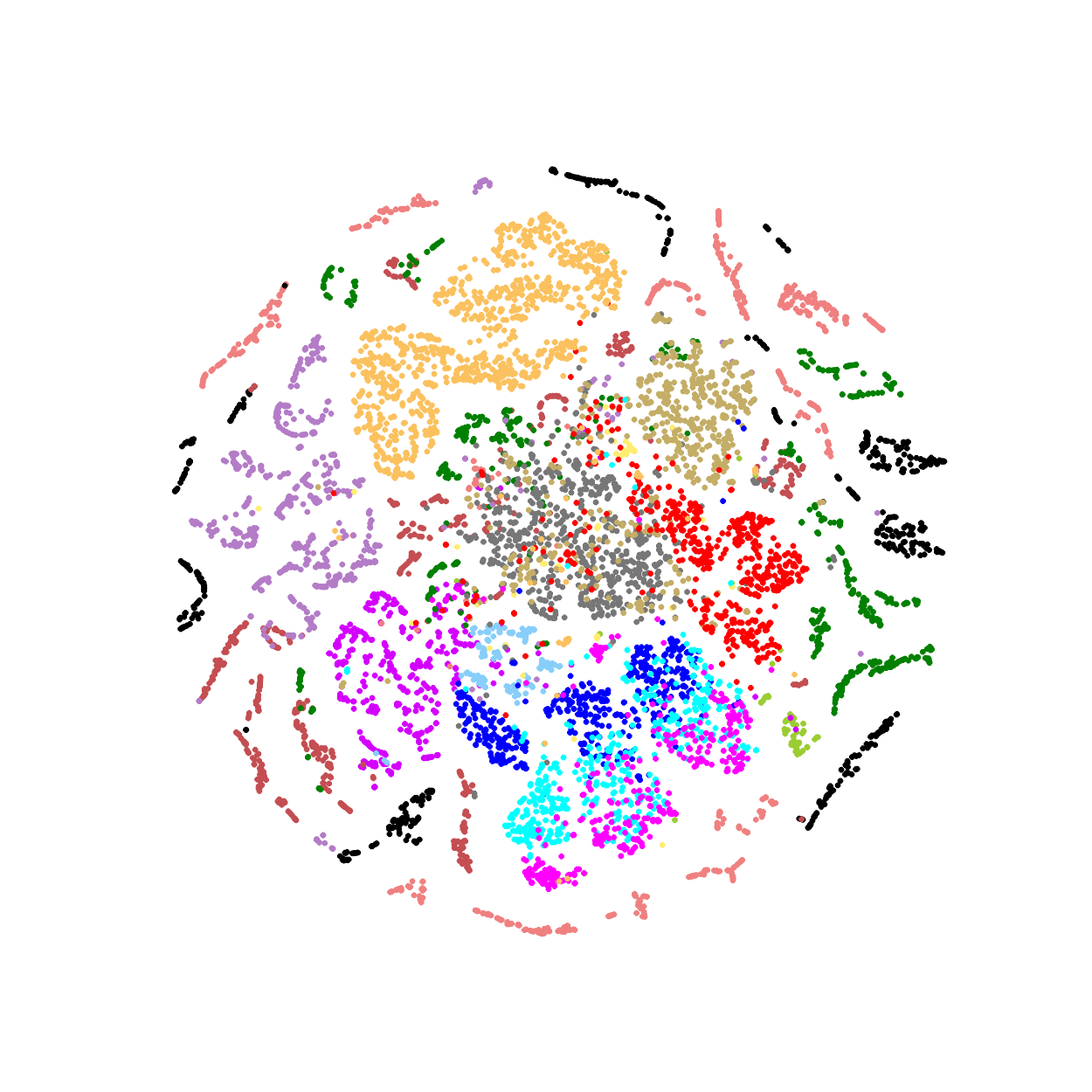}
        \caption{BPD(CNN) Redundant Features}
        \label{fig:bpd_ni}
    \end{subfigure}
    \caption{Feature visualisation: t-SNE plot of CNN features, activity features and redundant features on GOTOV Dataset. We use different colours to denote different categories.(best viewed in colour)}
    \label{fig:tsne-ab}
\end{figure}

Figure~\ref{fig:tsne-ab} illustrates the t-SNE plot for the original CNN feature~(Figure \ref{fig:cnn}), activity feature ~(Figure \ref{fig:bpd_ai}) and redundant features~(Figure \ref{fig:bpd_ni}). We can witness that the clusters of activity embeddings of BPD~(i.e., Figure \ref{fig:bpd_ai}) are more distinct and organised than those of CNN and redundant features~(i.e., Figure \ref{fig:cnn} and \ref{fig:bpd_ni}), and samples with the same activity class tend to group into the same cluster, i.e., smaller intra-class variability. 
For example,  ``Dish washing'', ``Vacuum Cleaning'', ``Stacking Shelves'' and ``Step''
seems more distinguishable in activity feature space than the other two. 
However, we also noticed that the redundant feature still contain some substantial activity patterns, which should be removed.  
One possible conjecture can be the limitation of this GOTOV dataset. Although it is a large dataset with 35 subjects, the population are older people (e.g., with age ranging from 61-73). The lack of diversity makes it challenging to remove the redundancy caused by personal factors completely. 

\subsection{Discussion and Limitation}
The proposed BPD framework aims to remove redundant features that do not contribute to the classification/recognition (in the training set), in order to reduce the intra-class variability for improved performance. For HAR scenarios, these redundant features correspond to the coupled effect of various covariate factors, e.g., the coupled effect of age, gender, weight, etc. and we expect performance gain by disentangling and removing these redundancies. However, current HAR datasets are normally limited due to population diversity, making it challenging to remove all the factors completely, and in Figure \ref{fig:bpd_ni}, we can still observe the activity patterns in the redundant features, suggesting the activity signals and the redundancy were not completely separated in the feature space. Nevertheless, our BPD framework can still reduce the effect of covariate factors, as suggested by the substantial performance gains on the 4 public datasets.

Although our framework can reduce the effect of covariate factors with improved performance,it remains unclear what these factors are. With additional metadata, a recent work GILE~\cite{2021GILE} attempted to disentangle human identity from activity signal, and this motivates us to explore further the attribute-oriented disentanglement frameworks. Although it may require additional meta-information for the disentanglement (between the attribute and the activity), it may provide a solution with higher interpretability.

Another major challenge is the cross-dataset evaluation.
Different datasets may be affected by more challenging external factors such as unpredictable wearing locations or unknown hardware settings. 
For example, accelerometer devices from various manufacturers may have different xyz orientations. 
Although our BPD can be used to reduce the intra-class variability to some extent at the person level, it is hard to generalise to unknown hardware setups. 
Some wearing location protocol or device calibration should be applied for the cross-dataset evaluation, which will be explored in the future.



\section{Conclusion}
The main focus of this work is developing a feature disentanglement method that can effectively disentangle the redundant information from the initial signal to reduce the intra-class variability and improve the performance of HAR models. Such a method could become a prerequisite for wider adoption of HAR models/algorithms in real-world applications. In this case, we proposed the Behaviour Pattern Disentanglement(BPD) scheme for sensor-based human activity recognition. Specifically, we first design a novel signal$\&$redundant feature disentanglement module that leverages the concept of adversary training to separate the activity feature and redundant feature from the initial representation. Then, we present a signal$\&$redundant feature dependency reduction module to reduce the correlation between two disentangled features so to improve the disentanglement. Finally, we present an adversarial training algorithm to ensure the proposed BPD framework can be trained properly. To evaluate the HAR models more thoroughly, we conducted extensive experiments on four public datasets. Experimental results suggested it can further improve the performance of existing DL approaches (e.g., CNN or DeepConvLSTM), making it a flexible solution for the HAR research community.


\bibliographystyle{ACM-Reference-Format}
\bibliography{bpd}

\begin{thebibliography}{58}


\ifx \showCODEN    \undefined \def \showCODEN     #1{\unskip}     \fi
\ifx \showDOI      \undefined \def \showDOI       #1{#1}\fi
\ifx \showISBNx    \undefined \def \showISBNx     #1{\unskip}     \fi
\ifx \showISBNxiii \undefined \def \showISBNxiii  #1{\unskip}     \fi
\ifx \showISSN     \undefined \def \showISSN      #1{\unskip}     \fi
\ifx \showLCCN     \undefined \def \showLCCN      #1{\unskip}     \fi
\ifx \shownote     \undefined \def \shownote      #1{#1}          \fi
\ifx \showarticletitle \undefined \def \showarticletitle #1{#1}   \fi
\ifx \showURL      \undefined \def \showURL       {\relax}        \fi
\providecommand\bibfield[2]{#2}
\providecommand\bibinfo[2]{#2}
\providecommand\natexlab[1]{#1}
\providecommand\showeprint[2][]{arXiv:#2}

\bibitem[\protect\citeauthoryear{Banos, Garcia, Holgado-Terriza, Damas,
  Pomares, Rojas, Saez, and Villalonga}{Banos et~al\mbox{.}}{2014}]%
        {MHEALTH}
\bibfield{author}{\bibinfo{person}{Oresti Banos}, \bibinfo{person}{Rafael
  Garcia}, \bibinfo{person}{Juan~A Holgado-Terriza}, \bibinfo{person}{Miguel
  Damas}, \bibinfo{person}{Hector Pomares}, \bibinfo{person}{Ignacio Rojas},
  \bibinfo{person}{Alejandro Saez}, {and} \bibinfo{person}{Claudia
  Villalonga}.} \bibinfo{year}{2014}\natexlab{}.
\newblock \showarticletitle{mHealthDroid: a novel framework for agile
  development of mobile health applications}. In
  \bibinfo{booktitle}{\emph{International workshop on ambient assisted
  living}}. Springer, \bibinfo{pages}{91--98}.
\newblock


\bibitem[\protect\citeauthoryear{Bao and Intille}{Bao and Intille}{2004}]%
        {bao2004activity}
\bibfield{author}{\bibinfo{person}{Ling Bao} {and} \bibinfo{person}{Stephen~S
  Intille}.} \bibinfo{year}{2004}\natexlab{}.
\newblock \showarticletitle{Activity recognition from user-annotated
  acceleration data}. In \bibinfo{booktitle}{\emph{International conference on
  pervasive computing}}. Springer, \bibinfo{pages}{1--17}.
\newblock


\bibitem[\protect\citeauthoryear{Barshan and Y{\"u}ksek}{Barshan and
  Y{\"u}ksek}{2014}]%
        {UCIDSADS}
\bibfield{author}{\bibinfo{person}{Billur Barshan} {and}
  \bibinfo{person}{Murat~Cihan Y{\"u}ksek}.} \bibinfo{year}{2014}\natexlab{}.
\newblock \showarticletitle{Recognizing daily and sports activities in two open
  source machine learning environments using body-worn sensor units}.
\newblock \bibinfo{journal}{\emph{Comput. J.}} \bibinfo{volume}{57},
  \bibinfo{number}{11} (\bibinfo{year}{2014}), \bibinfo{pages}{1649--1667}.
\newblock


\bibitem[\protect\citeauthoryear{Belghazi, Baratin, Rajeshwar, Ozair, Bengio,
  Courville, and Hjelm}{Belghazi et~al\mbox{.}}{2018}]%
        {belghazi2018mutual}
\bibfield{author}{\bibinfo{person}{Mohamed~Ishmael Belghazi},
  \bibinfo{person}{Aristide Baratin}, \bibinfo{person}{Sai Rajeshwar},
  \bibinfo{person}{Sherjil Ozair}, \bibinfo{person}{Yoshua Bengio},
  \bibinfo{person}{Aaron Courville}, {and} \bibinfo{person}{Devon Hjelm}.}
  \bibinfo{year}{2018}\natexlab{}.
\newblock \showarticletitle{Mutual information neural estimation}. In
  \bibinfo{booktitle}{\emph{International Conference on Machine Learning}}.
  PMLR, \bibinfo{pages}{531--540}.
\newblock


\bibitem[\protect\citeauthoryear{Bulling, Blanke, and Schiele}{Bulling
  et~al\mbox{.}}{2014}]%
        {bulling2014tutorial}
\bibfield{author}{\bibinfo{person}{Andreas Bulling}, \bibinfo{person}{Ulf
  Blanke}, {and} \bibinfo{person}{Bernt Schiele}.}
  \bibinfo{year}{2014}\natexlab{}.
\newblock \showarticletitle{A tutorial on human activity recognition using
  body-worn inertial sensors}.
\newblock \bibinfo{journal}{\emph{ACM Computing Surveys (CSUR)}}
  \bibinfo{volume}{46}, \bibinfo{number}{3} (\bibinfo{year}{2014}),
  \bibinfo{pages}{1--33}.
\newblock


\bibitem[\protect\citeauthoryear{Chakraborty, Alam, Dey, Chattopadhyay, and
  Mukhopadhyay}{Chakraborty et~al\mbox{.}}{2018}]%
        {chakraborty2018adversarial}
\bibfield{author}{\bibinfo{person}{Anirban Chakraborty},
  \bibinfo{person}{Manaar Alam}, \bibinfo{person}{Vishal Dey},
  \bibinfo{person}{Anupam Chattopadhyay}, {and} \bibinfo{person}{Debdeep
  Mukhopadhyay}.} \bibinfo{year}{2018}\natexlab{}.
\newblock \showarticletitle{Adversarial attacks and defences: A survey}.
\newblock \bibinfo{journal}{\emph{arXiv preprint arXiv:1810.00069}}
  (\bibinfo{year}{2018}).
\newblock


\bibitem[\protect\citeauthoryear{Chen, Duan, Houthooft, Schulman, Sutskever,
  and Abbeel}{Chen et~al\mbox{.}}{2016}]%
        {chen2016infogan}
\bibfield{author}{\bibinfo{person}{Xi Chen}, \bibinfo{person}{Yan Duan},
  \bibinfo{person}{Rein Houthooft}, \bibinfo{person}{John Schulman},
  \bibinfo{person}{Ilya Sutskever}, {and} \bibinfo{person}{Pieter Abbeel}.}
  \bibinfo{year}{2016}\natexlab{}.
\newblock \showarticletitle{Infogan: Interpretable representation learning by
  information maximizing generative adversarial nets}.
\newblock \bibinfo{journal}{\emph{arXiv preprint arXiv:1606.03657}}
  (\bibinfo{year}{2016}).
\newblock


\bibitem[\protect\citeauthoryear{Gao, Long, Guan, Basu, Baggaley, and
  Ploetz}{Gao et~al\mbox{.}}{2019}]%
        {babystroke}
\bibfield{author}{\bibinfo{person}{Yan Gao}, \bibinfo{person}{Yang Long},
  \bibinfo{person}{Yu Guan}, \bibinfo{person}{Anna Basu},
  \bibinfo{person}{Jessica Baggaley}, {and} \bibinfo{person}{Thomas Ploetz}.}
  \bibinfo{year}{2019}\natexlab{}.
\newblock \showarticletitle{Towards Reliable, Automated General Movement
  Assessment for Perinatal Stroke Screening in Infants Using Wearable
  Accelerometers}.
\newblock \bibinfo{journal}{\emph{Proc. ACM Interact. Mob. Wearable Ubiquitous
  Technol.}} \bibinfo{volume}{3}, \bibinfo{number}{1}, Article
  \bibinfo{articleno}{12} (\bibinfo{date}{March} \bibinfo{year}{2019}),
  \bibinfo{numpages}{22}~pages.
\newblock
\urldef\tempurl%
\url{https://doi.org/10.1145/3314399}
\showDOI{\tempurl}


\bibitem[\protect\citeauthoryear{Geirhos, Jacobsen, Michaelis, Zemel, Brendel,
  Bethge, and Wichmann}{Geirhos et~al\mbox{.}}{2020}]%
        {geirhos2020shortcut}
\bibfield{author}{\bibinfo{person}{Robert Geirhos},
  \bibinfo{person}{J{\"o}rn-Henrik Jacobsen}, \bibinfo{person}{Claudio
  Michaelis}, \bibinfo{person}{Richard Zemel}, \bibinfo{person}{Wieland
  Brendel}, \bibinfo{person}{Matthias Bethge}, {and} \bibinfo{person}{Felix~A
  Wichmann}.} \bibinfo{year}{2020}\natexlab{}.
\newblock \showarticletitle{Shortcut learning in deep neural networks}.
\newblock \bibinfo{journal}{\emph{Nature Machine Intelligence}}
  \bibinfo{volume}{2}, \bibinfo{number}{11} (\bibinfo{year}{2020}),
  \bibinfo{pages}{665--673}.
\newblock


\bibitem[\protect\citeauthoryear{Glorot and Bengio}{Glorot and Bengio}{2010}]%
        {glorot2010understanding}
\bibfield{author}{\bibinfo{person}{Xavier Glorot} {and} \bibinfo{person}{Yoshua
  Bengio}.} \bibinfo{year}{2010}\natexlab{}.
\newblock \showarticletitle{Understanding the difficulty of training deep
  feedforward neural networks}. In \bibinfo{booktitle}{\emph{Proceedings of the
  thirteenth international conference on artificial intelligence and
  statistics}}. JMLR Workshop and Conference Proceedings,
  \bibinfo{pages}{249--256}.
\newblock


\bibitem[\protect\citeauthoryear{Goodfellow, Pouget-Abadie, Mirza, Xu,
  Warde-Farley, Ozair, Courville, and Bengio}{Goodfellow et~al\mbox{.}}{2014}]%
        {goodfellow2014generative}
\bibfield{author}{\bibinfo{person}{Ian~J Goodfellow}, \bibinfo{person}{Jean
  Pouget-Abadie}, \bibinfo{person}{Mehdi Mirza}, \bibinfo{person}{Bing Xu},
  \bibinfo{person}{David Warde-Farley}, \bibinfo{person}{Sherjil Ozair},
  \bibinfo{person}{Aaron Courville}, {and} \bibinfo{person}{Yoshua Bengio}.}
  \bibinfo{year}{2014}\natexlab{}.
\newblock \showarticletitle{Generative adversarial networks}.
\newblock \bibinfo{journal}{\emph{arXiv preprint arXiv:1406.2661}}
  (\bibinfo{year}{2014}).
\newblock


\bibitem[\protect\citeauthoryear{Gopalan, Taheri, Turaga, and
  Chellappa}{Gopalan et~al\mbox{.}}{2012}]%
        {gopalan2012blur}
\bibfield{author}{\bibinfo{person}{Raghuraman Gopalan}, \bibinfo{person}{Sima
  Taheri}, \bibinfo{person}{Pavan Turaga}, {and} \bibinfo{person}{Rama
  Chellappa}.} \bibinfo{year}{2012}\natexlab{}.
\newblock \showarticletitle{A blur-robust descriptor with applications to face
  recognition}.
\newblock \bibinfo{journal}{\emph{IEEE transactions on pattern analysis and
  machine intelligence}} \bibinfo{volume}{34}, \bibinfo{number}{6}
  (\bibinfo{year}{2012}), \bibinfo{pages}{1220--1226}.
\newblock


\bibitem[\protect\citeauthoryear{Guan and Pl{\"o}tz}{Guan and
  Pl{\"o}tz}{2017}]%
        {2017ensembleslstm}
\bibfield{author}{\bibinfo{person}{Yu Guan} {and} \bibinfo{person}{Thomas
  Pl{\"o}tz}.} \bibinfo{year}{2017}\natexlab{}.
\newblock \showarticletitle{Ensembles of deep lstm learners for activity
  recognition using wearables}.
\newblock \bibinfo{journal}{\emph{Proceedings of the ACM on Interactive,
  Mobile, Wearable and Ubiquitous Technologies}} \bibinfo{volume}{1},
  \bibinfo{number}{2} (\bibinfo{year}{2017}), \bibinfo{pages}{1--28}.
\newblock


\bibitem[\protect\citeauthoryear{Guo, Chen, Peng, and Chen}{Guo
  et~al\mbox{.}}{2016}]%
        {guo2016wearable}
\bibfield{author}{\bibinfo{person}{Haodong Guo}, \bibinfo{person}{Ling Chen},
  \bibinfo{person}{Liangying Peng}, {and} \bibinfo{person}{Gencai Chen}.}
  \bibinfo{year}{2016}\natexlab{}.
\newblock \showarticletitle{Wearable sensor based multimodal human activity
  recognition exploiting the diversity of classifier ensemble}. In
  \bibinfo{booktitle}{\emph{Proceedings of the 2016 ACM International Joint
  Conference on Pervasive and Ubiquitous Computing}}.
  \bibinfo{pages}{1112--1123}.
\newblock


\bibitem[\protect\citeauthoryear{Hammerla, Halloran, and Pl{\"o}tz}{Hammerla
  et~al\mbox{.}}{2016a}]%
        {2016bLSTMS}
\bibfield{author}{\bibinfo{person}{Nils~Y Hammerla}, \bibinfo{person}{Shane
  Halloran}, {and} \bibinfo{person}{Thomas Pl{\"o}tz}.}
  \bibinfo{year}{2016}\natexlab{a}.
\newblock \showarticletitle{Deep, convolutional, and recurrent models for human
  activity recognition using wearables}.
\newblock \bibinfo{journal}{\emph{arXiv preprint arXiv:1604.08880}}
  (\bibinfo{year}{2016}).
\newblock


\bibitem[\protect\citeauthoryear{Hammerla, Halloran, and Pl{\"o}tz}{Hammerla
  et~al\mbox{.}}{2016b}]%
        {hammerla2016deep}
\bibfield{author}{\bibinfo{person}{Nils~Y Hammerla}, \bibinfo{person}{Shane
  Halloran}, {and} \bibinfo{person}{Thomas Pl{\"o}tz}.}
  \bibinfo{year}{2016}\natexlab{b}.
\newblock \showarticletitle{Deep, convolutional, and recurrent models for human
  activity recognition using wearables}.
\newblock \bibinfo{journal}{\emph{arXiv preprint arXiv:1604.08880}}
  (\bibinfo{year}{2016}).
\newblock


\bibitem[\protect\citeauthoryear{Higgins, Matthey, Pal, Burgess, Glorot,
  Botvinick, Mohamed, and Lerchner}{Higgins et~al\mbox{.}}{2016}]%
        {higgins2016beta}
\bibfield{author}{\bibinfo{person}{Irina Higgins}, \bibinfo{person}{Loic
  Matthey}, \bibinfo{person}{Arka Pal}, \bibinfo{person}{Christopher Burgess},
  \bibinfo{person}{Xavier Glorot}, \bibinfo{person}{Matthew Botvinick},
  \bibinfo{person}{Shakir Mohamed}, {and} \bibinfo{person}{Alexander
  Lerchner}.} \bibinfo{year}{2016}\natexlab{}.
\newblock \showarticletitle{beta-vae: Learning basic visual concepts with a
  constrained variational framework}.
\newblock  (\bibinfo{year}{2016}).
\newblock


\bibitem[\protect\citeauthoryear{Hochreiter and Schmidhuber}{Hochreiter and
  Schmidhuber}{1997}]%
        {hochreiter1997long}
\bibfield{author}{\bibinfo{person}{Sepp Hochreiter} {and}
  \bibinfo{person}{J{\"u}rgen Schmidhuber}.} \bibinfo{year}{1997}\natexlab{}.
\newblock \showarticletitle{Long short-term memory}.
\newblock \bibinfo{journal}{\emph{Neural computation}} \bibinfo{volume}{9},
  \bibinfo{number}{8} (\bibinfo{year}{1997}), \bibinfo{pages}{1735--1780}.
\newblock


\bibitem[\protect\citeauthoryear{Hu, Gao, Guan, Long, Lane, and Ploetz}{Hu
  et~al\mbox{.}}{2018a}]%
        {hu2018robust}
\bibfield{author}{\bibinfo{person}{BingZhang Hu}, \bibinfo{person}{Yan Gao},
  \bibinfo{person}{Yu Guan}, \bibinfo{person}{Yang Long},
  \bibinfo{person}{Nicholas Lane}, {and} \bibinfo{person}{Thomas Ploetz}.}
  \bibinfo{year}{2018}\natexlab{a}.
\newblock \showarticletitle{Robust cross-view gait identification with
  evidence: A discriminant gait gan (diggan) approach on 10000 people}.
\newblock \bibinfo{journal}{\emph{arXiv e-prints}} (\bibinfo{year}{2018}),
  \bibinfo{pages}{arXiv--1811}.
\newblock


\bibitem[\protect\citeauthoryear{Hu, Guan, Gao, Long, Lane, and Ploetz}{Hu
  et~al\mbox{.}}{2020}]%
        {hu2020robust}
\bibfield{author}{\bibinfo{person}{BingZhang Hu}, \bibinfo{person}{Yu Guan},
  \bibinfo{person}{Yan Gao}, \bibinfo{person}{Yang Long},
  \bibinfo{person}{Nicholas Lane}, {and} \bibinfo{person}{Thomas Ploetz}.}
  \bibinfo{year}{2020}\natexlab{}.
\newblock \bibinfo{title}{Robust Cross-View Gait Recognition with Evidence: A
  Discriminant Gait GAN (DiGGAN) Approach}.
\newblock
\newblock
\showeprint[arxiv]{1811.10493}~[cs.CV]


\bibitem[\protect\citeauthoryear{Hu, Zheng, and Shao}{Hu
  et~al\mbox{.}}{2018b}]%
        {hu2018dual}
\bibfield{author}{\bibinfo{person}{BingZhang Hu}, \bibinfo{person}{Feng Zheng},
  {and} \bibinfo{person}{Ling Shao}.} \bibinfo{year}{2018}\natexlab{b}.
\newblock \showarticletitle{Dual-Reference Face Retrieval}. In
  \bibinfo{booktitle}{\emph{Proceedings of the AAAI Conference on Artificial
  Intelligence}}, Vol.~\bibinfo{volume}{32}.
\newblock


\bibitem[\protect\citeauthoryear{Jing, Yang, Feng, Ye, Yu, and Song}{Jing
  et~al\mbox{.}}{2019}]%
        {jing2019neural}
\bibfield{author}{\bibinfo{person}{Yongcheng Jing}, \bibinfo{person}{Yezhou
  Yang}, \bibinfo{person}{Zunlei Feng}, \bibinfo{person}{Jingwen Ye},
  \bibinfo{person}{Yizhou Yu}, {and} \bibinfo{person}{Mingli Song}.}
  \bibinfo{year}{2019}\natexlab{}.
\newblock \showarticletitle{Neural style transfer: A review}.
\newblock \bibinfo{journal}{\emph{IEEE transactions on visualization and
  computer graphics}} \bibinfo{volume}{26}, \bibinfo{number}{11}
  (\bibinfo{year}{2019}), \bibinfo{pages}{3365--3385}.
\newblock


\bibitem[\protect\citeauthoryear{Khan, Nicholson, and Pl\"{o}tz}{Khan
  et~al\mbox{.}}{2017}]%
        {Cricket}
\bibfield{author}{\bibinfo{person}{Aftab Khan}, \bibinfo{person}{James
  Nicholson}, {and} \bibinfo{person}{Thomas Pl\"{o}tz}.}
  \bibinfo{year}{2017}\natexlab{}.
\newblock \showarticletitle{Activity Recognition for Quality Assessment of
  Batting Shots in Cricket Using a Hierarchical Representation}.
\newblock \bibinfo{journal}{\emph{Proc. ACM Interact. Mob. Wearable Ubiquitous
  Technol.}} \bibinfo{volume}{1}, \bibinfo{number}{3}, Article
  \bibinfo{articleno}{62} (\bibinfo{date}{Sept.} \bibinfo{year}{2017}),
  \bibinfo{numpages}{31}~pages.
\newblock
\urldef\tempurl%
\url{https://doi.org/10.1145/3130927}
\showDOI{\tempurl}


\bibitem[\protect\citeauthoryear{Kingma and Ba}{Kingma and Ba}{2014}]%
        {kingma2014adam}
\bibfield{author}{\bibinfo{person}{Diederik~P Kingma} {and}
  \bibinfo{person}{Jimmy Ba}.} \bibinfo{year}{2014}\natexlab{}.
\newblock \showarticletitle{Adam: A method for stochastic optimization}.
\newblock \bibinfo{journal}{\emph{arXiv preprint arXiv:1412.6980}}
  (\bibinfo{year}{2014}).
\newblock


\bibitem[\protect\citeauthoryear{Kingma and Welling}{Kingma and
  Welling}{2013}]%
        {kingma2013auto}
\bibfield{author}{\bibinfo{person}{Diederik~P Kingma} {and}
  \bibinfo{person}{Max Welling}.} \bibinfo{year}{2013}\natexlab{}.
\newblock \showarticletitle{Auto-encoding variational bayes}.
\newblock \bibinfo{journal}{\emph{arXiv preprint arXiv:1312.6114}}
  (\bibinfo{year}{2013}).
\newblock


\bibitem[\protect\citeauthoryear{Kinney and Atwal}{Kinney and Atwal}{2014}]%
        {kinney2014equitability}
\bibfield{author}{\bibinfo{person}{Justin~B Kinney} {and}
  \bibinfo{person}{Gurinder~S Atwal}.} \bibinfo{year}{2014}\natexlab{}.
\newblock \showarticletitle{Equitability, mutual information, and the maximal
  information coefficient}.
\newblock \bibinfo{journal}{\emph{Proceedings of the National Academy of
  Sciences}} \bibinfo{volume}{111}, \bibinfo{number}{9} (\bibinfo{year}{2014}),
  \bibinfo{pages}{3354--3359}.
\newblock


\bibitem[\protect\citeauthoryear{Kwapisz, Weiss, and Moore}{Kwapisz
  et~al\mbox{.}}{2011}]%
        {kwapisz2011activity}
\bibfield{author}{\bibinfo{person}{Jennifer~R Kwapisz}, \bibinfo{person}{Gary~M
  Weiss}, {and} \bibinfo{person}{Samuel~A Moore}.}
  \bibinfo{year}{2011}\natexlab{}.
\newblock \showarticletitle{Activity recognition using cell phone
  accelerometers}.
\newblock \bibinfo{journal}{\emph{ACM SigKDD Explorations Newsletter}}
  \bibinfo{volume}{12}, \bibinfo{number}{2} (\bibinfo{year}{2011}),
  \bibinfo{pages}{74--82}.
\newblock


\bibitem[\protect\citeauthoryear{LeCun, Bengio, et~al\mbox{.}}{LeCun
  et~al\mbox{.}}{1995}]%
        {lecun1995convolutional}
\bibfield{author}{\bibinfo{person}{Yann LeCun}, \bibinfo{person}{Yoshua
  Bengio}, {et~al\mbox{.}}} \bibinfo{year}{1995}\natexlab{}.
\newblock \showarticletitle{Convolutional networks for images, speech, and time
  series}.
\newblock \bibinfo{journal}{\emph{The handbook of brain theory and neural
  networks}} \bibinfo{volume}{3361}, \bibinfo{number}{10}
  (\bibinfo{year}{1995}), \bibinfo{pages}{1995}.
\newblock


\bibitem[\protect\citeauthoryear{Liu, Liu, Yeh, and Wang}{Liu
  et~al\mbox{.}}{2018}]%
        {liu2018unified}
\bibfield{author}{\bibinfo{person}{Alexander~H Liu}, \bibinfo{person}{Yen-Cheng
  Liu}, \bibinfo{person}{Yu-Ying Yeh}, {and} \bibinfo{person}{Yu-Chiang~Frank
  Wang}.} \bibinfo{year}{2018}\natexlab{}.
\newblock \showarticletitle{A unified feature disentangler for multi-domain
  image translation and manipulation}.
\newblock \bibinfo{journal}{\emph{arXiv preprint arXiv:1809.01361}}
  (\bibinfo{year}{2018}).
\newblock


\bibitem[\protect\citeauthoryear{Locatello, Abbati, Rainforth, Bauer,
  Sch{\"o}lkopf, and Bachem}{Locatello et~al\mbox{.}}{2019a}]%
        {locatello2019fairness}
\bibfield{author}{\bibinfo{person}{Francesco Locatello},
  \bibinfo{person}{Gabriele Abbati}, \bibinfo{person}{Tom Rainforth},
  \bibinfo{person}{Stefan Bauer}, \bibinfo{person}{Bernhard Sch{\"o}lkopf},
  {and} \bibinfo{person}{Olivier Bachem}.} \bibinfo{year}{2019}\natexlab{a}.
\newblock \showarticletitle{On the fairness of disentangled representations}.
\newblock \bibinfo{journal}{\emph{arXiv preprint arXiv:1905.13662}}
  (\bibinfo{year}{2019}).
\newblock


\bibitem[\protect\citeauthoryear{Locatello, Bauer, Lucic, Raetsch, Gelly,
  Sch{\"o}lkopf, and Bachem}{Locatello et~al\mbox{.}}{2019b}]%
        {locatello2019challenging}
\bibfield{author}{\bibinfo{person}{Francesco Locatello},
  \bibinfo{person}{Stefan Bauer}, \bibinfo{person}{Mario Lucic},
  \bibinfo{person}{Gunnar Raetsch}, \bibinfo{person}{Sylvain Gelly},
  \bibinfo{person}{Bernhard Sch{\"o}lkopf}, {and} \bibinfo{person}{Olivier
  Bachem}.} \bibinfo{year}{2019}\natexlab{b}.
\newblock \showarticletitle{Challenging common assumptions in the unsupervised
  learning of disentangled representations}. In
  \bibinfo{booktitle}{\emph{international conference on machine learning}}.
  PMLR, \bibinfo{pages}{4114--4124}.
\newblock


\bibitem[\protect\citeauthoryear{Locatello, Tschannen, Bauer, R{\"a}tsch,
  Sch{\"o}lkopf, and Bachem}{Locatello et~al\mbox{.}}{2019c}]%
        {locatello2019disentangling}
\bibfield{author}{\bibinfo{person}{Francesco Locatello},
  \bibinfo{person}{Michael Tschannen}, \bibinfo{person}{Stefan Bauer},
  \bibinfo{person}{Gunnar R{\"a}tsch}, \bibinfo{person}{Bernhard
  Sch{\"o}lkopf}, {and} \bibinfo{person}{Olivier Bachem}.}
  \bibinfo{year}{2019}\natexlab{c}.
\newblock \showarticletitle{Disentangling factors of variation using few
  labels}.
\newblock \bibinfo{journal}{\emph{arXiv preprint arXiv:1905.01258}}
  (\bibinfo{year}{2019}).
\newblock


\bibitem[\protect\citeauthoryear{Lowe}{Lowe}{2004}]%
        {lowe2004distinctive}
\bibfield{author}{\bibinfo{person}{David~G Lowe}.}
  \bibinfo{year}{2004}\natexlab{}.
\newblock \showarticletitle{Distinctive image features from scale-invariant
  keypoints}.
\newblock \bibinfo{journal}{\emph{International journal of computer vision}}
  \bibinfo{volume}{60}, \bibinfo{number}{2} (\bibinfo{year}{2004}),
  \bibinfo{pages}{91--110}.
\newblock


\bibitem[\protect\citeauthoryear{Mathieu, Zhao, Sprechmann, Ramesh, and
  LeCun}{Mathieu et~al\mbox{.}}{2016}]%
        {mathieu2016disentangling}
\bibfield{author}{\bibinfo{person}{Michael Mathieu}, \bibinfo{person}{Junbo
  Zhao}, \bibinfo{person}{Pablo Sprechmann}, \bibinfo{person}{Aditya Ramesh},
  {and} \bibinfo{person}{Yann LeCun}.} \bibinfo{year}{2016}\natexlab{}.
\newblock \showarticletitle{Disentangling factors of variation in deep
  representations using adversarial training}.
\newblock \bibinfo{journal}{\emph{arXiv preprint arXiv:1611.03383}}
  (\bibinfo{year}{2016}).
\newblock


\bibitem[\protect\citeauthoryear{Murahari and Pl{\"o}tz}{Murahari and
  Pl{\"o}tz}{2018}]%
        {2018attentionlstm}
\bibfield{author}{\bibinfo{person}{Vishvak~S Murahari} {and}
  \bibinfo{person}{Thomas Pl{\"o}tz}.} \bibinfo{year}{2018}\natexlab{}.
\newblock \showarticletitle{On attention models for human activity
  recognition}. In \bibinfo{booktitle}{\emph{Proceedings of the 2018 ACM
  International Symposium on Wearable Computers}}. \bibinfo{pages}{100--103}.
\newblock


\bibitem[\protect\citeauthoryear{Odena, Olah, and Shlens}{Odena
  et~al\mbox{.}}{2017}]%
        {odena2017conditional}
\bibfield{author}{\bibinfo{person}{Augustus Odena},
  \bibinfo{person}{Christopher Olah}, {and} \bibinfo{person}{Jonathon Shlens}.}
  \bibinfo{year}{2017}\natexlab{}.
\newblock \showarticletitle{Conditional image synthesis with auxiliary
  classifier gans}. In \bibinfo{booktitle}{\emph{International conference on
  machine learning}}. PMLR, \bibinfo{pages}{2642--2651}.
\newblock


\bibitem[\protect\citeauthoryear{Ord{\'o}{\~n}ez and Roggen}{Ord{\'o}{\~n}ez
  and Roggen}{2016}]%
        {2016DeepConvLSTM}
\bibfield{author}{\bibinfo{person}{Francisco~Javier Ord{\'o}{\~n}ez} {and}
  \bibinfo{person}{Daniel Roggen}.} \bibinfo{year}{2016}\natexlab{}.
\newblock \showarticletitle{Deep convolutional and lstm recurrent neural
  networks for multimodal wearable activity recognition}.
\newblock \bibinfo{journal}{\emph{Sensors}} \bibinfo{volume}{16},
  \bibinfo{number}{1} (\bibinfo{year}{2016}), \bibinfo{pages}{115}.
\newblock


\bibitem[\protect\citeauthoryear{Ozair, Lynch, Bengio, Oord, Levine, and
  Sermanet}{Ozair et~al\mbox{.}}{2019}]%
        {ozair2019wasserstein}
\bibfield{author}{\bibinfo{person}{Sherjil Ozair}, \bibinfo{person}{Corey
  Lynch}, \bibinfo{person}{Yoshua Bengio}, \bibinfo{person}{Aaron van~den
  Oord}, \bibinfo{person}{Sergey Levine}, {and} \bibinfo{person}{Pierre
  Sermanet}.} \bibinfo{year}{2019}\natexlab{}.
\newblock \showarticletitle{Wasserstein dependency measure for representation
  learning}.
\newblock \bibinfo{journal}{\emph{arXiv preprint arXiv:1903.11780}}
  (\bibinfo{year}{2019}).
\newblock


\bibitem[\protect\citeauthoryear{Pal}{Pal}{2005}]%
        {pal2005random}
\bibfield{author}{\bibinfo{person}{Mahesh Pal}.}
  \bibinfo{year}{2005}\natexlab{}.
\newblock \showarticletitle{Random forest classifier for remote sensing
  classification}.
\newblock \bibinfo{journal}{\emph{International journal of remote sensing}}
  \bibinfo{volume}{26}, \bibinfo{number}{1} (\bibinfo{year}{2005}),
  \bibinfo{pages}{217--222}.
\newblock


\bibitem[\protect\citeauthoryear{Paraschiakos, Cachucho, Moed, van Heemst,
  Mooijaart, Slagboom, Knobbe, and Beekman}{Paraschiakos et~al\mbox{.}}{2020}]%
        {GOTOV}
\bibfield{author}{\bibinfo{person}{Stylianos Paraschiakos},
  \bibinfo{person}{Ricardo Cachucho}, \bibinfo{person}{Matthijs Moed},
  \bibinfo{person}{Diana van Heemst}, \bibinfo{person}{Simon Mooijaart},
  \bibinfo{person}{Eline~P Slagboom}, \bibinfo{person}{Arno Knobbe}, {and}
  \bibinfo{person}{Marian Beekman}.} \bibinfo{year}{2020}\natexlab{}.
\newblock \showarticletitle{Activity recognition using wearable sensors for
  tracking the elderly}.
\newblock \bibinfo{journal}{\emph{User Modeling and User-Adapted Interaction}}
  \bibinfo{volume}{30}, \bibinfo{number}{3} (\bibinfo{year}{2020}),
  \bibinfo{pages}{567--605}.
\newblock


\bibitem[\protect\citeauthoryear{Paszke, Gross, Massa, Lerer, Bradbury, Chanan,
  Killeen, Lin, Gimelshein, Antiga, Desmaison, Kopf, Yang, DeVito, Raison,
  Tejani, Chilamkurthy, Steiner, Fang, Bai, and Chintala}{Paszke
  et~al\mbox{.}}{2019}]%
        {pytorch}
\bibfield{author}{\bibinfo{person}{Adam Paszke}, \bibinfo{person}{Sam Gross},
  \bibinfo{person}{Francisco Massa}, \bibinfo{person}{Adam Lerer},
  \bibinfo{person}{James Bradbury}, \bibinfo{person}{Gregory Chanan},
  \bibinfo{person}{Trevor Killeen}, \bibinfo{person}{Zeming Lin},
  \bibinfo{person}{Natalia Gimelshein}, \bibinfo{person}{Luca Antiga},
  \bibinfo{person}{Alban Desmaison}, \bibinfo{person}{Andreas Kopf},
  \bibinfo{person}{Edward Yang}, \bibinfo{person}{Zachary DeVito},
  \bibinfo{person}{Martin Raison}, \bibinfo{person}{Alykhan Tejani},
  \bibinfo{person}{Sasank Chilamkurthy}, \bibinfo{person}{Benoit Steiner},
  \bibinfo{person}{Lu Fang}, \bibinfo{person}{Junjie Bai}, {and}
  \bibinfo{person}{Soumith Chintala}.} \bibinfo{year}{2019}\natexlab{}.
\newblock \showarticletitle{PyTorch: An Imperative Style, High-Performance Deep
  Learning Library}.
\newblock In \bibinfo{booktitle}{\emph{Advances in Neural Information
  Processing Systems 32}}, \bibfield{editor}{\bibinfo{person}{H.~Wallach},
  \bibinfo{person}{H.~Larochelle}, \bibinfo{person}{A.~Beygelzimer},
  \bibinfo{person}{F.~d\textquotesingle Alch\'{e}-Buc},
  \bibinfo{person}{E.~Fox}, {and} \bibinfo{person}{R.~Garnett}} (Eds.).
  \bibinfo{publisher}{Curran Associates, Inc.}, \bibinfo{pages}{8024--8035}.
\newblock
\urldef\tempurl%
\url{http://papers.neurips.cc/paper/9015-pytorch-an-imperative-style-high-performance-deep-learning-library.pdf}
\showURL{%
\tempurl}


\bibitem[\protect\citeauthoryear{Pl\"{o}tz, Hammerla, Rozga, Reavis, Call, and
  Abowd}{Pl\"{o}tz et~al\mbox{.}}{2012}]%
        {Autism}
\bibfield{author}{\bibinfo{person}{Thomas Pl\"{o}tz}, \bibinfo{person}{Nils~Y.
  Hammerla}, \bibinfo{person}{Agata Rozga}, \bibinfo{person}{Andrea Reavis},
  \bibinfo{person}{Nathan Call}, {and} \bibinfo{person}{Gregory~D. Abowd}.}
  \bibinfo{year}{2012}\natexlab{}.
\newblock \showarticletitle{Automatic Assessment of Problem Behavior in
  Individuals with Developmental Disabilities}. In
  \bibinfo{booktitle}{\emph{Proceedings of the 2012 ACM Conference on
  Ubiquitous Computing}} (Pittsburgh, Pennsylvania)
  \emph{(\bibinfo{series}{UbiComp '12})}. \bibinfo{publisher}{Association for
  Computing Machinery}, \bibinfo{address}{New York, NY, USA},
  \bibinfo{pages}{391–400}.
\newblock
\showISBNx{9781450312240}
\urldef\tempurl%
\url{https://doi.org/10.1145/2370216.2370276}
\showDOI{\tempurl}


\bibitem[\protect\citeauthoryear{Qian, Su, Wen, Jha, Li, Guan, Puthal, James,
  Yang, Zomaya, et~al\mbox{.}}{Qian et~al\mbox{.}}{2020}]%
        {qian2020orchestrating}
\bibfield{author}{\bibinfo{person}{Bin Qian}, \bibinfo{person}{Jie Su},
  \bibinfo{person}{Zhenyu Wen}, \bibinfo{person}{Devki~Nandan Jha},
  \bibinfo{person}{Yinhao Li}, \bibinfo{person}{Yu Guan},
  \bibinfo{person}{Deepak Puthal}, \bibinfo{person}{Philip James},
  \bibinfo{person}{Renyu Yang}, \bibinfo{person}{Albert~Y Zomaya},
  {et~al\mbox{.}}} \bibinfo{year}{2020}\natexlab{}.
\newblock \showarticletitle{Orchestrating the development lifecycle of machine
  learning-based iot applications: A taxonomy and survey}.
\newblock \bibinfo{journal}{\emph{ACM Computing Surveys (CSUR)}}
  \bibinfo{volume}{53}, \bibinfo{number}{4} (\bibinfo{year}{2020}),
  \bibinfo{pages}{1--47}.
\newblock


\bibitem[\protect\citeauthoryear{Qian, Pan, Da, and Miao}{Qian
  et~al\mbox{.}}{2019a}]%
        {qian2019novel}
\bibfield{author}{\bibinfo{person}{Hangwei Qian}, \bibinfo{person}{Sinno~Jialin
  Pan}, \bibinfo{person}{Bingshui Da}, {and} \bibinfo{person}{Chunyan Miao}.}
  \bibinfo{year}{2019}\natexlab{a}.
\newblock \showarticletitle{A Novel Distribution-Embedded Neural Network for
  Sensor-Based Activity Recognition.}. In \bibinfo{booktitle}{\emph{IJCAI}}.
  \bibinfo{pages}{5614--5620}.
\newblock


\bibitem[\protect\citeauthoryear{Qian, Pan, Da, and Miao}{Qian
  et~al\mbox{.}}{2019b}]%
        {2019DDNN}
\bibfield{author}{\bibinfo{person}{Hangwei Qian}, \bibinfo{person}{Sinno~Jialin
  Pan}, \bibinfo{person}{Bingshui Da}, {and} \bibinfo{person}{Chunyan Miao}.}
  \bibinfo{year}{2019}\natexlab{b}.
\newblock \showarticletitle{A novel distribution-embedded neural network for
  sensor-based activity recognition}.
\newblock  (\bibinfo{year}{2019}).
\newblock


\bibitem[\protect\citeauthoryear{Qian, Pan, and Miao}{Qian
  et~al\mbox{.}}{2021}]%
        {2021GILE}
\bibfield{author}{\bibinfo{person}{Hangwei Qian}, \bibinfo{person}{Sinno~Jialin
  Pan}, {and} \bibinfo{person}{Chunyan Miao}.} \bibinfo{year}{2021}\natexlab{}.
\newblock \showarticletitle{Latent Independent Excitation for Generalizable
  Sensor-based Cross-Person Activity Recognition}. In
  \bibinfo{booktitle}{\emph{Thirty-Fifth {AAAI} Conference on Artificial
  Intelligence, {AAAI} 2021, Thirty-Third Conference on Innovative Applications
  of Artificial Intelligence, {IAAI} 2021, The Eleventh Symposium on
  Educational Advances in Artificial Intelligence, {EAAI} 2021, Virtual Event,
  February 2-9, 2021}}. \bibinfo{publisher}{{AAAI} Press},
  \bibinfo{pages}{11921--11929}.
\newblock
\urldef\tempurl%
\url{https://ojs.aaai.org/index.php/AAAI/article/view/17416}
\showURL{%
\tempurl}


\bibitem[\protect\citeauthoryear{Reiss and Stricker}{Reiss and
  Stricker}{2012}]%
        {PAMAP2}
\bibfield{author}{\bibinfo{person}{Attila Reiss} {and} \bibinfo{person}{Didier
  Stricker}.} \bibinfo{year}{2012}\natexlab{}.
\newblock \showarticletitle{Introducing a new benchmarked dataset for activity
  monitoring}. In \bibinfo{booktitle}{\emph{2012 16th International Symposium
  on Wearable Computers}}. IEEE, \bibinfo{pages}{108--109}.
\newblock


\bibitem[\protect\citeauthoryear{Ridgeway and Mozer}{Ridgeway and
  Mozer}{2018}]%
        {ridgeway2018learning}
\bibfield{author}{\bibinfo{person}{Karl Ridgeway} {and}
  \bibinfo{person}{Michael~C Mozer}.} \bibinfo{year}{2018}\natexlab{}.
\newblock \showarticletitle{Learning deep disentangled embeddings with the
  f-statistic loss}.
\newblock \bibinfo{journal}{\emph{arXiv preprint arXiv:1802.05312}}
  (\bibinfo{year}{2018}).
\newblock


\bibitem[\protect\citeauthoryear{Sanchez, Serrurier, and Ortner}{Sanchez
  et~al\mbox{.}}{2020}]%
        {sanchez2020learning}
\bibfield{author}{\bibinfo{person}{Eduardo~Hugo Sanchez},
  \bibinfo{person}{Mathieu Serrurier}, {and} \bibinfo{person}{Mathias Ortner}.}
  \bibinfo{year}{2020}\natexlab{}.
\newblock \showarticletitle{Learning disentangled representations via mutual
  information estimation}. In \bibinfo{booktitle}{\emph{European Conference on
  Computer Vision}}. Springer, \bibinfo{pages}{205--221}.
\newblock


\bibitem[\protect\citeauthoryear{Sang, Xia, and Hansen}{Sang
  et~al\mbox{.}}{2020}]%
        {sang2020deaan}
\bibfield{author}{\bibinfo{person}{Mufan Sang}, \bibinfo{person}{Wei Xia},
  {and} \bibinfo{person}{John~HL Hansen}.} \bibinfo{year}{2020}\natexlab{}.
\newblock \showarticletitle{DEAAN: Disentangled Embedding and Adversarial
  Adaptation Network for Robust Speaker Representation Learning}.
\newblock \bibinfo{journal}{\emph{arXiv preprint arXiv:2012.06896}}
  (\bibinfo{year}{2020}).
\newblock


\bibitem[\protect\citeauthoryear{Schmidt, Beigl, and Gellersen}{Schmidt
  et~al\mbox{.}}{1999}]%
        {schmidt1999there}
\bibfield{author}{\bibinfo{person}{Albrecht Schmidt}, \bibinfo{person}{Michael
  Beigl}, {and} \bibinfo{person}{Hans-W Gellersen}.}
  \bibinfo{year}{1999}\natexlab{}.
\newblock \showarticletitle{There is more to context than location}.
\newblock \bibinfo{journal}{\emph{Computers \& Graphics}} \bibinfo{volume}{23},
  \bibinfo{number}{6} (\bibinfo{year}{1999}), \bibinfo{pages}{893--901}.
\newblock


\bibitem[\protect\citeauthoryear{Szegedy, Zaremba, Sutskever, Bruna, Erhan,
  Goodfellow, and Fergus}{Szegedy et~al\mbox{.}}{2013}]%
        {szegedy2013intriguing}
\bibfield{author}{\bibinfo{person}{Christian Szegedy},
  \bibinfo{person}{Wojciech Zaremba}, \bibinfo{person}{Ilya Sutskever},
  \bibinfo{person}{Joan Bruna}, \bibinfo{person}{Dumitru Erhan},
  \bibinfo{person}{Ian Goodfellow}, {and} \bibinfo{person}{Rob Fergus}.}
  \bibinfo{year}{2013}\natexlab{}.
\newblock \showarticletitle{Intriguing properties of neural networks}.
\newblock \bibinfo{journal}{\emph{arXiv preprint arXiv:1312.6199}}
  (\bibinfo{year}{2013}).
\newblock


\bibitem[\protect\citeauthoryear{van Steenkiste, Locatello, Schmidhuber, and
  Bachem}{van Steenkiste et~al\mbox{.}}{2019}]%
        {van2019disentangled}
\bibfield{author}{\bibinfo{person}{Sjoerd van Steenkiste},
  \bibinfo{person}{Francesco Locatello}, \bibinfo{person}{J{\"u}rgen
  Schmidhuber}, {and} \bibinfo{person}{Olivier Bachem}.}
  \bibinfo{year}{2019}\natexlab{}.
\newblock \showarticletitle{Are disentangled representations helpful for
  abstract visual reasoning?}
\newblock \bibinfo{journal}{\emph{arXiv preprint arXiv:1905.12506}}
  (\bibinfo{year}{2019}).
\newblock


\bibitem[\protect\citeauthoryear{Wang, Hu, Long, and Guan}{Wang
  et~al\mbox{.}}{2019}]%
        {wang2019order}
\bibfield{author}{\bibinfo{person}{Junyan Wang}, \bibinfo{person}{Bingzhang
  Hu}, \bibinfo{person}{Yang Long}, {and} \bibinfo{person}{Yu Guan}.}
  \bibinfo{year}{2019}\natexlab{}.
\newblock \showarticletitle{Order matters: Shuffling sequence generation for
  video prediction}.
\newblock \bibinfo{journal}{\emph{arXiv preprint arXiv:1907.08845}}
  (\bibinfo{year}{2019}).
\newblock


\bibitem[\protect\citeauthoryear{Xiao, Li, Zhu, He, Liu, and Song}{Xiao
  et~al\mbox{.}}{2018}]%
        {xiao2018generating}
\bibfield{author}{\bibinfo{person}{Chaowei Xiao}, \bibinfo{person}{Bo Li},
  \bibinfo{person}{Jun-Yan Zhu}, \bibinfo{person}{Warren He},
  \bibinfo{person}{Mingyan Liu}, {and} \bibinfo{person}{Dawn Song}.}
  \bibinfo{year}{2018}\natexlab{}.
\newblock \showarticletitle{Generating adversarial examples with adversarial
  networks}.
\newblock \bibinfo{journal}{\emph{arXiv preprint arXiv:1801.02610}}
  (\bibinfo{year}{2018}).
\newblock


\bibitem[\protect\citeauthoryear{Yang, Nguyen, San, Li, and Krishnaswamy}{Yang
  et~al\mbox{.}}{2015}]%
        {2015MCCNN}
\bibfield{author}{\bibinfo{person}{Jianbo Yang}, \bibinfo{person}{Minh~Nhut
  Nguyen}, \bibinfo{person}{Phyo~Phyo San}, \bibinfo{person}{Xiaoli Li}, {and}
  \bibinfo{person}{Shonali Krishnaswamy}.} \bibinfo{year}{2015}\natexlab{}.
\newblock \showarticletitle{Deep convolutional neural networks on multichannel
  time series for human activity recognition.}. In
  \bibinfo{booktitle}{\emph{Ijcai}}, Vol.~\bibinfo{volume}{15}. Buenos Aires,
  Argentina, \bibinfo{pages}{3995--4001}.
\newblock


\bibitem[\protect\citeauthoryear{Zhai, Perez-Pozuelo, Clifton, Palotti, and
  Guan}{Zhai et~al\mbox{.}}{2020}]%
        {bing2020}
\bibfield{author}{\bibinfo{person}{Bing Zhai}, \bibinfo{person}{Ignacio
  Perez-Pozuelo}, \bibinfo{person}{Emma A.~D. Clifton}, \bibinfo{person}{Joao
  Palotti}, {and} \bibinfo{person}{Yu Guan}.} \bibinfo{year}{2020}\natexlab{}.
\newblock \showarticletitle{Making Sense of Sleep: Multimodal Sleep Stage
  Classification in a Large, Diverse Population Using Movement and Cardiac
  Sensing}.
\newblock \bibinfo{journal}{\emph{Proc. ACM Interact. Mob. Wearable Ubiquitous
  Technol.}} \bibinfo{volume}{4}, \bibinfo{number}{2}, Article
  \bibinfo{articleno}{67} (\bibinfo{date}{June} \bibinfo{year}{2020}),
  \bibinfo{numpages}{33}~pages.
\newblock
\urldef\tempurl%
\url{https://doi.org/10.1145/3397325}
\showDOI{\tempurl}


\bibitem[\protect\citeauthoryear{Zhang, Zhang, Zhang, Liu, and Khurshid}{Zhang
  et~al\mbox{.}}{2018}]%
        {zhang2018deeproad}
\bibfield{author}{\bibinfo{person}{Mengshi Zhang}, \bibinfo{person}{Yuqun
  Zhang}, \bibinfo{person}{Lingming Zhang}, \bibinfo{person}{Cong Liu}, {and}
  \bibinfo{person}{Sarfraz Khurshid}.} \bibinfo{year}{2018}\natexlab{}.
\newblock \showarticletitle{DeepRoad: GAN-based metamorphic testing and input
  validation framework for autonomous driving systems}. In
  \bibinfo{booktitle}{\emph{2018 33rd IEEE/ACM International Conference on
  Automated Software Engineering (ASE)}}. IEEE, \bibinfo{pages}{132--142}.
\newblock


\end{thebibliography}

\newpage
\appendix
\section{GOTOV dataset result}
\label{appdix_gotov}
\begin{table}[h]
\caption{Mean F1-score for each subject on the GOTOV dataset (in leave-one-subject-out CV setting)}
\label{appdix:result}
\begin{tabular}{c|c|cccccc}
\hline
Dataset                 & Subject & CNN    & DeepConvLSTM & beta-VAE &\hspace{0.5cm} GILE  \hspace{0.5cm} & \begin{tabular}[c]{@{}c@{}}BPD\\ (CNN)\end{tabular} & \begin{tabular}[c]{@{}c@{}}BPD\\ (DeepConvLSTM)\end{tabular} \\ \hline
                         & 1       & 0.6852 & 0.7156       & 0.6200   & 0.6401 & \textbf{0.7541} & 0.7493                                 \\
                         & 2       & 0.7585 & 0.7241       & 0.7064   & 0.7120 & \textbf{0.7730} & 0.7287                                 \\
                         & 3       & 0.6770 & 0.6432       & 0.6221   & 0.6373 & \textbf{0.7357} & 0.6469                                 \\
                         & 4       & 0.7480 & 0.6993       & 0.5751   & 0.6931 & \textbf{0.7497} & 0.7263                                 \\
                         & 5       & 0.8295 & 0.8157       & 0.6757   & 0.7832 & \textbf{0.8378} & 0.8271                                 \\
                         & 6       & 0.6608 & 0.6569       & 0.5572   & 0.6212 & \textbf{0.6742}          & 0.6670                                        \\
                         & 7       & 0.6581 & 0.7125       & 0.5845   & 0.6101 & 0.7194          & \textbf{0.7391}                        \\
                         & 8       & 0.7843 & 0.8176       & 0.6750   & 0.7532 & 0.8118          & \textbf{0.8378}                        \\
                         & 9       & 0.8945 & 0.8997       & 0.7365   & 0.8106 & 0.8968          & \textbf{0.9018}                        \\
                         & 10      & 0.6584 & 0.6527       & 0.5780   & 0.5701 & \textbf{0.6621} & 0.6537                                 \\
                         & 11      & 0.7950 & 0.7915       & 0.6649   & 0.7432 & 0.7984          & \textbf{0.8164}                        \\
                         & 12      & 0.6802 & 0.6816       & 0.5479   & 0.6330 & \textbf{0.7458} & 0.6894          \\
                         & 13      & 0.7515 & 0.6415       & 0.5382   & 0.6179 & \textbf{0.7880} & 0.6725                                 \\
                         & 14      & 0.5257 & 0.5047       & 0.3700   & 0.3841 & \textbf{0.6243} & 0.5231                                 \\
                         & 15      & 0.7852 & 0.7746       & 0.6751   & 0.7597 & 0.7946          & \textbf{0.7961}                        \\
                         & 16      & 0.5837 & 0.5936       & 0.5311   & 0.5332 & \textbf{0.6537} & 0.6210                                 \\
                         & 17      & 0.6124 & 0.6055       & 0.5328   & 0.5421 & \textbf{0.6476} & 0.6319                                 \\
                         & 18      & 0.5664 & 0.5102       & 0.4213   & 0.4979 & \textbf{0.5709} & 0.5305          \\
                         & 19      & 0.6900 & 0.6533       & 0.6192   & 0.6127 & \textbf{0.7530} & 0.6825                                 \\
                         & 20      & 0.8163 & 0.8190       & 0.7243   & 0.7904 & \textbf{0.8725} & 0.8219                                 \\
                         & 21      & 0.7902 & 0.7756       & 0.5745   & 0.6920 & \textbf{0.8037}          & 0.7869                                       \\
                         & 22      & 0.4930 & 0.4564       & 0.3693   & 0.3988 & \textbf{0.5122} & 0.4826                                 \\
                         & 23      & 0.5523 & 0.5647       & 0.5030   & 0.4988 & 0.5540          &  \textbf{0.5841} \\
                         & 24      & 0.6332 & 0.6250       & 0.6078   & 0.6320 & \textbf{0.6643} & 0.6374         \\
                         & 25      & 0.5303 & 0.4571       & 0.3908   & 0.4450 & \textbf{0.5340} & 0.4842                                 \\
                         & 26      & 0.4463 & 0.4204       & 0.3347   & 0.4102 & \textbf{0.4813} & 0.4455                                 \\
                         & 27      & 0.8107 & 0.7798       & 0.7157   & 0.7322 & \textbf{0.8261} & 0.7910          \\
                         & 28      & 0.6305 & 0.6046       & 0.5876   & 0.5886 & \textbf{0.6529} & 0.6377                                 \\
                         & 29      & 0.4893 & 0.4578       & 0.3921   & 0.4172 & \textbf{0.4951} & 0.4692          \\
                         & 30      & 0.6301 & 0.6405       & 0.5354   & 0.6078 & 0.6799          & \textbf{0.6840}                        \\
                         & 31      & 0.5527 & 0.5454       & 0.4941   & 0.5132 & \textbf{0.5929} & 0.5595                                 \\
                         & 32      & 0.7328 & 0.7065       & 0.6184   & 0.6932 & \textbf{0.7620}          & 0.7388                                       \\
                         & 33      & 0.6779 & 0.6427       & 0.5948   & 0.6321 & \textbf{0.6986} & 0.6505                                 \\
                         & 34      & 0.6307 & 0.6169       & 0.5212   & 0.6039 & \textbf{0.6490} & 0.6453                                 \\
                         & 35      & 0.4982 & 0.5163       & 0.3670   & 0.4821 & \textbf{0.5670} & 0.5639                                 \\ \cline{2-8}
\multirow{-36}{*}{GOTOV} & Avg.    & 0.6645 & 0.6492       & 0.5589   & 0.6083 & \textbf{0.6953}          & 0.6692 \\ \hline                               
\end{tabular}
\end{table}

\end{document}